\documentclass[10pt,journal,compsoc]{IEEEtran} 
\usepackage{microtype}
\usepackage{graphicx}
\usepackage{subfigure}
\usepackage{pifont}
\usepackage{amsfonts}
\usepackage{booktabs} 
\usepackage{bbm}
\usepackage{bm}
\usepackage{soul}
\usepackage{adjustbox} 
\usepackage{hyperref}    
\usepackage{color} 
\usepackage[dvipsnames]{xcolor}
\usepackage{enumitem,kantlipsum}
\usepackage[sort&compress,numbers]{natbib}

\usepackage{amssymb}
\usepackage{comment} 
\usepackage{amsthm}
\usepackage{dsfont}
\usepackage{amsmath}
\usepackage[justification=centering]{caption}

\newcommand{\ie}{\textit{i.e.}}
\newcommand{\eg}{\textit{e.g.}}
\newcommand{\stt}{\textit{s.t.}}

\newcommand{\cmark}{\ding{51}}%
\newcommand{\xmark}{\ding{55}}%
\newcommand{\RR}{\mathbb{R}}
\newcommand{\cA}{{\mathcal{A}}}
\newcommand{\cH}{{\mathcal{H}}}
\newcommand{\cT}{{\mathcal{T}}}
\newcommand{\cD}{{\mathcal{D}}}
\newcommand{\cI}{{\mathcal{I}}}
\newcommand{\cL}{{\mathcal{L}}}
\newcommand{\cM}{{\mathcal{M}}}

\newcommand{\cX}{{\mathcal{X}}}
\newcommand{\cone}{{\mathbbm{1}}}
\newcommand{\cS}{{\mathcal{S}}}

\newcommand{\cR}{{\mathcal{R}}}
\newcommand{\cG}{{\mathcal{G}}}
\newcommand{\cE}{{\mathds{E}}}
\newcommand{\cF}{{\mathcal{F}}}

\newcommand{\vq}{{\bm{q}}}

\newcommand{\vW}{{\mathbf{w}}}

\newcommand{\vphi}{{\bm{\phi}}}
\newcommand{\vpsi}{{\bm{\psi}}}
\newcommand{\tb}{\textbf}
\newcommand{\judycom}[1]{{\color{black}{#1}}}
\newcommand{\modify}[1]{{\color{black}{#1}}}

\DeclareMathOperator*{\argmax}{arg\,max}

\newtheorem*{theorem*}{Theorem}

\newtheorem*{assumption*}{Assumption}

\newtheorem*{policyTransfer*}{Policy Transfer}

\newtheorem{remark}{Remark}
\newtheorem*{remark*}{Remark}
 
\begin{document}
\title{Transfer Learning in Deep Reinforcement Learning: A Survey}
\author{Zhuangdi~Zhu,~
        Kaixiang~Lin,
        Anil K. Jain,
        and~Jiayu~Zhou
\IEEEcompsocitemizethanks{\IEEEcompsocthanksitem Zhuangdi Zhu, Anil K. Jain, and Jiayu Zhou are with the Department
of Computer Science and Engineering, Michigan State University, East Lansing,
MI, 48824.~\protect 
E-mail: \{zhuzhuan, jain, jiayuz\}@msu.edu

\IEEEcompsocthanksitem Kaixiang Lin is with the Amazon Alexa AI. \protect
E-mail: lkxcarson@gmail.com
}
}



\IEEEtitleabstractindextext{%
\begin{abstract}
    %
    Reinforcement learning is a learning paradigm for solving sequential decision-making problems. 
    Recent years have witnessed remarkable progress in reinforcement learning upon the fast development of deep neural networks. Along with the promising prospects of reinforcement learning in numerous domains such as robotics and game-playing, transfer learning has arisen to tackle various challenges faced by reinforcement learning, by transferring knowledge from external expertise to
    facilitate the efficiency and effectiveness of the learning process.
    In this survey, we systematically investigate the recent progress of transfer learning approaches in the context of deep reinforcement learning. Specifically, we provide a framework for categorizing the state-of-the-art transfer learning approaches, under which we analyze their goals, methodologies, compatible reinforcement learning backbones, and practical applications. We also draw connections between transfer learning and other relevant topics from the reinforcement learning perspective and explore their potential challenges that await future research progress.
\end{abstract}

}

\maketitle
\IEEEdisplaynontitleabstractindextext 
\IEEEpeerreviewmaketitle

\newpage
\IEEEraisesectionheading{\section{Introduction}\label{sec:introduction}}

\IEEEPARstart{R}einforcement Learning (RL) is an effective framework to solve sequential decision-making tasks, where a learning agent interacts with the environment to improve its performance through trial and error~\citep{sutton2018reinforcement}.
Originated from cybernetics and thriving in computer science, RL has been widely applied to tackle challenging tasks which were previously intractable.
Traditional RL algorithms were mostly designed for tabular cases, which provide principled solutions to simple tasks but face difficulties when handling highly complex domains, \eg~tasks with 3D environments.
With the recent advances in deep learning research, the combination of RL and deep neural networks is developed to address challenging tasks.
The combination of deep learning with RL is hence referred to as \textit{\textbf{Deep Reinforcement Learning}} (DRL)~\cite{arulkumaran2017brief}, which learns powerful function approximators using deep neural networks to address complicated domains.
DRL has achieved notable success in applications such as robotics control~\cite{levine2016end,levine2018learning} and game playing~\cite{bellemare2013arcade}.
It also thrives in domains such as health informatics~\cite{kosorok2015adaptive}, electricity networks~\cite{glavic2017reinforcement}, intelligent transportation systems\cite{el2013multiagent,wei2018intellilight}, to name just a few. 

Besides its remarkable advancement, RL still faces intriguing difficulties
induced by the \textit{exploration-exploitation} dilemma~\cite{sutton2018reinforcement}.
Specifically, for practical RL problems, the environment dynamics are usually unknown, and the agent cannot exploit knowledge about the environment until enough interaction experiences are collected via exploration. 
Due to the partial observability, sparse feedbacks, and the high complexity of state and action spaces, acquiring sufficient interaction samples can be prohibitive or even incur safety concerns for domains such as automatic-driving and health informatics.
The abovementioned challenges have motivated various efforts to improve the current RL procedure. 
As a result, \emph{\textbf{transfer learning}} (TL), or equivalently referred as \emph{\textbf{knowledge transfer}}, which is a technique to utilize external expertise to benefit the learning process of the target domain, becomes a crucial topic in RL.

While TL techniques have been extensively studied in  \textit{supervised learning}~\cite{pan2009survey}, it is still an emerging topic for RL. 
Transfer learning can be more complicated for RL, in that the knowledge needs to transfer in the context of a {Markov Decision Process}. 
Moreover, due to the delicate components of the Markov decision process, expert knowledge may take different forms that need to transfer in different ways.      
\modify{
	Noticing that previous efforts on summarizing TL in the RL domain did not cover research of the last decade~\citep{taylor2009transfer, lazaric2012transfer}, during which time considerate TL breakthroughs have been achieved empowered with deep learning techniques.
	Hence, in this survey, we make a comprehensive investigation of the latest TL approaches in RL.
}
%

%
\judycom{
The contributions of our survey are multifold:
1) we investigated up-to-date research involving {new DRL backbones} and TL algorithms over the recent decade.  
To the best of our knowledge, this survey is the first attempt to survey TL approaches in the context of \textbf{\emph{deep}} reinforcement learning.
We reviewed TL methods that can tackle more evolved RL tasks,
%
and also studied new TL schemes that are not deeply discussed by prior literatures, such as {representation disentanglement}~(Sec \ref{sec:invariant}) and {policy distillation} (Sec \ref{sec:policyTransfer}).
2) We provided systematic categorizations that cover a broader and deeper view of TL developments in DRL.
Our main analysis is anchored on a fundamental question, \ie\ \emph{what is the transferred knowledge in RL}, following which we conducted more refined analysis.
Most TL strategies, including those discussed in prior surveys are well suited in our categorization framework.
3) Reflecting on the developments of TL methods in DRL, we brought new thoughts on its future directions, including how to do \emph{reasoning} over miscellaneous knowledge forms and how to \emph{leverage} knowledge in more efficient and principled manner.
We also pointed out the prominent applications of TL for DRL and its opportunities to thrive in the future era of AGI.
}

The rest of this survey is organized as follows: 
In Section \ref{sec:bg} we introduce RL preliminaries, including the recent key development based on deep neural networks.
Next, we discuss the definition of TL in the context of RL and its relevant topics (Section \ref{sec:relatedTopics}).
In Section \ref{sec:evaluation}, we provide a framework to categorize TL approaches from multiple perspectives, analyze their fundamental differences, 
and summarize their evaluation metrics (Section \ref{sec:metrics}).
In Section \ref{sec:approaches}, we elaborate on different TL approaches in the context of DRL, organized by the format of transferred knowledge, such as \textit{reward shaping} (Section \ref{sec:rs}), \textit{learning from demonstrations} (Section \ref{sec:lfd}), or \textit{learning from teacher policies} (Section \ref{sec:policyTransfer}). We also investigate TL approaches by the way that knowledge transfer occurs, such as inter-task mapping (Section \ref{sec:mapping}), or learning transferrable representations (Section \ref{sec:invariant}), etc. 
We discuss contemporary  applications of TL in the context of DRL in Section \ref{sec:applications} and provide some future perspectives and open questions in Section \ref{sec:futurePerspectives}.


\section{Deep Reinforcement Learning and Transfer Learning} \label{sec:bg}

\subsection{Reinforcement Learning Basics}
\textbf{\emph{Markov Decision Process:}}
A typical RL problem can be considered as training an agent to interact with an environment that follows a {\textit{Markov Decision Process} (MPD)}~\cite{bellman1957markovian}. 
%
The agent starts with an initial \textit{state} and performs an \emph{action} accordingly, which yields a \emph{reward} to guide the agent actions. Once the action is taken, the MDP transits to the next state by following the underlying \emph{transition dynamics} of the MDP. 
The agent accumulates the time-\emph{discounted} rewards along with its interactions. 
A subsequence of interactions is referred to as an \emph{episode}.
The above-mentioned components in an MDP can be represented using a tuple, \ie~$ \cM=(\mu_0, \cS, \cA, \cT, \gamma, \cR$), in which: 
\vspace{-0.05in}
\begin{itemize}[leftmargin=*]%

\item $\mu_0$ is the set of \textit{\textbf{initial states}}. 
\item $\cS$ is the \textit{\textbf{state}} space. 
\item $\cA$ is the \textit{\textbf{action}} space.
%
\item $\cT$: $S \times A \times S \to \RR $ is the \textit{\textbf{transition probability distribution}}, where $\cT(s' |s,a)$ specifies the probability of the state transitioning to $s'$ upon taking action $a$ from state $s$. 
\item $\cR: S \times A \times S \to \RR$ is the \textit{\textbf{reward distribution}}, where $\cR(s, a, s')$ is the reward that an agent can get by taking action $a$ from state $s$ with the next state being $s'$.
\item $\gamma$ is a discounted factor, with $\gamma \in (0, 1]$.  
%
\end{itemize}

A RL agent behaves in $\cM$ by following its policy $\pi$, which is a mapping from states to actions: 
$ 
    \pi: \cS \to \cA 
$ 
. For a stochastic policy $\pi$, $\pi(a|s)$ denotes the probability of taking action $a$ from state $s$.
Given an MDP $\cM$ and a policy $\pi$, one can derive a \textit{\tb{value function}} $ V_\cM^{\pi}(s)$, which is defined over the state space:
   $ V_\cM^{\pi}(s) = \cE \left[ r_0 + \gamma r_1 + \gamma ^ 2 r_2 + \dots ; \pi, s \right],$
where $r_i = \cR(s_i, a_i, s_{i+1})$ is the reward that an agent receives by taking action $a_i$ in the $i$-th state $s_i$, and the next state transits to $s_{i+1}$.
The expectation $\cE$ is taken over ${s_0 \sim \mu_0, a_i \sim \pi(\cdot|s_i), s_{i+1} \sim \cT(\cdot| s_i,a_i) }$. 
The value function estimates the \textit{quality} of being in state $s$, by evaluating the expected rewards that an agent can get from $s$ following policy $\pi$. 
%
Similar to the value function, a policy also carries a \tb{$Q$-function}, which estimates the quality of taking action $a$ from state $s$:
  $  Q_\cM^{\pi}(s, a) = \cE_{s' \sim \cT(\cdot| s, a)}  \left[ \cR(s, a, s') + \gamma V_\cM^{\pi}(s') \right].$
 %
\vspace{0.1in} 
\judycom{
\noindent \emph{\textbf{Reinforcement Learning Goals:}}
    Standard RL aims  to learn an optimal policy $\pi_{\cM}^{*}$ with the optimal value and $Q$-function, \stt\ $\forall s \in \cS,\pi_\cM^{*} (s) = \underset{ a\in A} {\arg \max}~~{Q_\cM^*(s,a)},$ where $Q_\cM^*(s,a)= \underset{\pi} {\sup} ~~ Q_\cM^{\pi}(s,a)$. The learning objective can be reduced as maximizing the expected return: 
    \begin{align*}
    J(\pi) := \cE_{(s,a) \sim \mu^\pi(s,a)} [ \sum_{t} \gamma^t r_t],
    \vspace{-0.1in}
    \end{align*}
    where $\mu^\pi(s,a)$ is the \emph{stationary state-action distribution} induced by $\pi$~\cite{bellemare2017distributional}. 

    Built upon recent progress of DRL, some literature has extended the RL objective to achieving miscellaneous goals under different conditions, referred to as \textbf{\emph{Goal-Conditional RL}} (GCRL).
    In GCRL, the agent policy $\pi(\cdot|s, g)$ is dependent not only on state observations $s$ but also the goal $g$ being optimized. 
    %
    Each individual goal $g \sim \cG$  can be differentiated by its reward function $r(s_t, a_t, g)$, hence the objective for GCRL becomes maximizing the expected return over the distribution of goals: $J(\pi) := \cE_{(s_t, a_t) \sim \mu^\pi, g\sim \cG} \left[ \sum_t \gamma^t r(s, a, g)\right]$ ~\cite{liu2022goal}.
    A prototype example of GCRL can be maze locomotion tasks, where the learning goals are manifested as desired locations in the maze~\cite{florensa2018automatic}. 
}

\vspace{0.1in}
\judycom{
\noindent\emph{\textbf{Episodic vs. Non-episodic Reinforcement Learning:}}
In episodic RL, the agent performs in finite episodes of length $H$, and will be \emph{reset} to an initial state $\in \mu_0$ upon the episode ends~\cite{sutton2018reinforcement}. 
Whereas in non-episodic RL, the learning agent continuously interacts with the MDP without any state reset~\cite{xu2020reinforcement}.
To encompass the episodic concept in infinite MDPs, episodic RL tasks usually assume the existence of a set of \textit{absorbing states} $\cS_0$, which indicates the {termination} of episodic tasks~\cite{yu2022surprising,kostrikov2018discriminator}, and any action taken upon an absorbing state will only transit to itself with zero rewards.
}

\subsection{Reinforcement Learning Algorithms} \label{sec:rlAlgorithms}
%
\judycom{
There are two major methods to conduct RL: \textbf{\emph{Model-Based}} and \textbf{\emph{Model-Free}}.
In \emph{model-based} RL, 
a learned or provided model of the MDP is used for policy learning.
In \emph{model-free} RL, optimal policy is learned without modeling the transition dynamics or reward functions. 
In this section, we start introducing RL techniques from a \textbf{\emph{model-free}} perspective, due to its relatively simplicity, which also provides foundations for many model-based methods.
}

\textbf{\emph{Prediction and Control}}: an RL problem can be disassembled into two subtasks: \emph{prediction} and \emph{control}~\cite{sutton2018reinforcement}. 
In the \emph{prediction} phase, the quality of the current policy is being evaluated.
%
In the \emph{control} phase or the \emph{policy improvement} phase,  the learning policy is adjusted based on evaluation results from the \emph{prediction} step.
Policies can be improved by iterating through these two steps, known as \emph{policy iteration}.

\modify{For \emph{model-free} policy iterations, the target policy is optimized without requiring knowledge of the MDP transition dynamics.}
Traditional model-free RL includes \textbf{\emph{Monte-Carlo}} methods, which estimates the value of each state using \emph{samples of episodes} starting from that state.
Monte-Carlo methods can be \emph{on-policy} if the samples are collected by following the target policy,
or \emph{off-policy} if the episodic samples are collected by following a \emph{behavior} policy that is different from the target policy.

\textbf{\emph{Temporal Difference (TD) Learning}} is an alternative to Monte-Carlo  for solving the \emph{prediction} problem. 
The key idea behind TD-learning is to learn the state quality function by \emph{bootstrapping}.
%
It can also be extended to solve the \emph{control} problem so that both value function and policy can get improved simultaneously.
%
Examples of \emph{on-policy} TD-learning algorithms include~\emph{SARSA} \cite{rummery1994line}, \emph{Expected SARSA}~\cite{van2009theoretical}, \emph{Actor-Critic} \cite{konda2000actor}, and its deep neural network extension called \emph{A3C} \cite{mnih2016asynchronous}.
The \emph{off-policy} TD-learning approaches include SAC~\cite{haarnoja2018soft} for continuous state-action spaces, and {$Q$-learning}~\cite{watkins1992q} for discrete state-action spaces, along with its variants built on deep-neural networks, such as DQN~\cite{mnih2015human}, Double-DQN~\cite{mnih2015human}, Rainbow~\cite{hessel2018rainbow}, etc.
TD-learning approaches focus more on estimating the state-action value functions.

\textbf{\emph{Policy Gradient}}, on the other hand, is a mechanism that emphasizes on direct optimization of a parameterizable policy. Traditional policy-gradient approaches include \emph{REINFORCE}~\cite{williams1992simple}.
Recent years have witnessed the joint presence of TD-learning and policy-gradient approaches.
%
Representative algorithms along this line include \emph{Trust region policy optimization (TRPO)}~\cite{schulman2015trust}, \emph{Proximal Policy optimization (PPO)}~\cite{schulman2017proximal}, \emph{Deterministic policy gradient (DPG)}~\cite{silver2014deterministic} and its extensions such as \emph{DDPG}~\cite{lillicrap2015continuous} and \emph{Twin Delayed DDPG}~\cite{fujimoto2018addressing}.

\judycom{
Unlike model-free methods that learn purely from trial-and-error, \textbf{\emph{Model-Based RL}} (MBRL) explicitly learns the transition dynamics or cost functions of the environment. 
The dynamics model can sometimes be treated as a \textbf{\emph{black-box}} for better \emph{sampling-based planning}. 
Representative examples include the \emph{Monte-Carlo} method dubbed \emph{random shooting}~\cite{nagabandi2018neural} and its {cross-entropy} method (CEM) variants~\cite{botev2013cross,chua2018deep}.
The modeled dynamics can also facilitate learning with data generation~\cite{sutton1990integrated} and value estimation~\cite{feinberg2018model}. 
For MBRL with \textbf{\emph{white-box}} modeling, 
the transition models become differentiable and can facilitate planning with direct gradient propogation. 
Methods along this line include \emph{differential planning}  for policy gradient~\cite{levine2013guided} and action sequences search~\cite{bharadhwaj2020model}, and value gradient methods~\cite{deisenroth2011pilco,gal2016improving}.
%
%
One advantage of MBRL is its higher sample efficiency than model-free RL,
although it can be challenging for complex domains, where it is usually more difficult to learn the dynamics than learning a policy. 
}

\subsection{Transfer Learning in the Context of Reinforcement Learning} \label{sec:TLinRL}
\begin{remark}
    Without losing clarify, for the rest of this survey, we refer to MDPs, domains, and tasks equivalently.
\end{remark}

\begin{remark} 
    \textbf{\emph{[Transfer Learning in the Context of RL]}} Given a set of \textbf{source} domains $\bm{\cM}_s = \{\cM_s | \cM_s \in \bm{\cM}_s \}$ and a \textbf{target} domain $\cM_t$, \emph{Transfer Learning} aims to learn an optimal policy $\pi^*$ for the target domain, by leveraging exterior information $\cI_s$ from $\bm{\cM}_s$ as well as interior information $\cI_t$ from $\cM_t$:  
\begin{align*} 
        &~ \pi^* = \argmax_\pi \cE_{s \sim \mu_0^{t}, a \sim \pi}[Q^\pi_{\cM}(s, a)],
\end{align*}
 where ${\pi}=\phi(\cI_s \sim \bm{\cM}_s, \cI_t \sim \cM_t):\cS^t \to \cA^t$ is a policy learned for the target domain $\cM_t$ based on information from both $\cI_t$ and $\cI_s$.
\end{remark}
In the above definition, we use $\phi(\cI)$ to denote the learned policy based on information $\cI$, which is usually approximated with deep neural networks in DRL.
For the simplistic case, knowledge can transfer between two agents within the same domain, resulting in $|\bm{\cM}_s|= 1$, and $\cM_s = \cM_t$. 
One can consider regular RL without TL as a special case of the above definition, by treating $\cI_s=\emptyset$, so that a policy $\pi$ is learned purely on the feedback provided by the target domain,  $i.e.~{\pi}=\phi(\cI_t)$.

\subsection{Related Topics} \label{sec:relatedTopics}
In addition to TL, other efforts have been made to benefit RL by leveraging different forms of supervision.
In this section, we briefly discuss other techniques that are relevant to TL by analyzing the differences and connections between transfer learning and these relevant techniques, which we hope can further clarify the scope of this survey.

\modify{
\emph{\textbf{Continual Learning}} is the ability of sequentially learning multiple tasks that are temporally or spatially related, without forgetting the previously acquired knowledge.
%
Continual Learning is a specialized yet more challenging scenario of TL, in that the learned knowledge needs to be transferred along a sequence of dynamically-changing tasks that cannot be foreseen, rather than learning a fixed group of tasks.
Hence, different from most TL methods discussed in this survey, the ability of \emph{automatic task detection} and \emph{avoiding catastrophic forgetting} is usually indispensable in continual learning~\cite{lampert2009learning}.
}

\modify{
\textbf{\emph{Hierarchical RL}} has been proposed to resolve complex real-world tasks.
Different from traditional RL, for hierarchical RL, the action space is grouped into different granularities to form higher-level macro actions.
Accordingly, the learning task is also decomposed into hierarchically dependent sub-goals.
Well-known hierarchical RL frameworks include  \emph{Feudal learning} \cite{dayan1993feudal}, \emph{Options framework}\cite{sutton1999between}, \emph{Hierarchical Abstract Machines} \cite{parr1998reinforcement}, and \emph{MAXQ} \cite{dietterich2000hierarchical}.
Given the higher-level abstraction on tasks, actions, and state spaces, 
hierarchical RL can facilitate knowledge transfer across similar domains.
}

\judycom{
    \textbf{\emph{Multi-task RL}} learns an agent with generalized skills across various tasks, hence it can solve MDPs randomly sampled from a fixed yet unknown distribution \cite{lazaric2010bayesian}. 
    A larger concept of {multi-task learning}  also incorporates multi-task \textit{supervised} learning and \textit{unsupervised} learning~\cite{zhang2021survey}.
    Multi-task learning is naturally related to TL, in that the learned skills, typically manifested as representations, need to be effectively shared among domains.
    Many TL techniques later discussed in this survey can be readily applied to solve multi-task RL scenarios, such as policy distillation \cite{teh2017distral}, and representation sharing \cite{parisotto2015actor}.
    One notable challenges in multi-task learning is \textit{negative transfer}, which is induced by the irrelevance or conflicting property for learned tasks.
    Hence, some recent work in multi-task RL focused on a trade-off between sharing and individualizing function modules~\cite{devin2017learning,andreas2017modular,yang2020multi}.
}

\judycom{
    \textbf{\emph{Generalization in RL}} refers to the ability of learning agents to adapt to \emph{unseen} domains.
    Generalization is a crucial property for RL to achieve, especially when classical RL assumes identical training and inference MDPs, whereas the real world is constantly changing.
    Generalization in RL is considered more challenging than in supervised learning due to the non-stationarity of MDPs, where the latter has provided inspirations for the former \cite{hospedales2021meta}. 
    \textbf{\emph{Meta-learning}} is an effective direction towards generalization, which also draws close connections to TL.
    Some TL techniques discussed in this survey are actually designed for meta-RL.
    However, meta-learning is particularly focused on the learning methods that lead to \emph{fast adaptation} to {unseen} domains, 
    whereas TL is a broader concept and covers scenarios where the target environment can be (partially) observable.
    To tackle unseen tasks in RL, some meta-RL methods focused on training MDPs generation~\cite{jia2022improving} and variations estimation~\cite{ding2022generalizing}.
    We refer readers to \cite{kirk2023survey} for a more focused survey on meta RL. 
}

\vspace{-0.1in}
\section{\judycom{Analyzing Transfer Learning}} \label{sec:evaluation} \vspace{-0.05in}
In this section, we discuss TL approaches in RL from different angles. We also use a prototype to illustrate the potential variants residing in knowledge transfer among domains, then summarize important metrics for TL evaluation.

\vspace{-0.1in}
\subsection{Categorization of Transfer Learning Approaches} \label{sec:categorize} \vspace{-0.05in}
TL approaches can be organized by answering the following key questions:

\begin{enumerate}[leftmargin=*]

    \item \noindent \textbf{\emph{What knowledge is transferred}}: 
    Knowledge from the source domain can take different forms, such as expert experiences~\cite{kim2013learning}, 
    the action probability distribution of an expert policy~\cite{czarnecki2019distilling}, 
    or even a potential function that estimates the quality of demonstrations in the target MDP~\cite{ng1999policy}.
    The divergence in \textit{representations} and \textit{granularities} of knowledge fundamentally influences how TL is performed.
   The \textit{quality} of the transferred knowledge, \eg~whether it comes from an oracle ~\cite{goodfellow2014generative} or a suboptimal teacher~\cite{zhu2020learning} also affects the way TL methods are designed.

    \item \noindent \textbf{\emph{What RL frameworks fit the TL approach:}}
    We can rephrase this question into other forms, e.g., \textit{is the TL approach policy-agnostic, or only applicable to certain RL backbones, such as the Temporal Difference (TD) methods?}
    Answers to this question are closely related to the representaion of knowledge.
    For example, transferring knowledge from expert demonstrations are usually policy-agnostic (see Section~\ref{sec:lfd}), while policy distillation, to be discussed in Section~\ref{sec:policyTransfer}, may not be suitable for DQN backbone which does not explicitly learn a policy function.

    \item \noindent \textbf{\emph{What is the difference between the source and the target domain}}: 
    %
    Some TL approaches fit where the source domain $\cM_s$ and the target domain $\cM_t$ are equivalent,
    whereas others are designed to transfer knowledge between different domains.
    %
    %
    For example, in video gaming tasks where observations are RGB pixels, $\cM_s$ and $\cM_t$ may share the same action space ($\cA$) but differs in their observation spaces ($\cS$).
    For  goal-conditioned RL~\cite{schaul2015universal}, the two domains may differ only by the reward distribution: $\cR_s \neq \cR_t$. 
    %

    \item \noindent \textbf{\emph{What information is available in the target domain:}}~
    While knowledge from source domains is usually accessible, it can be prohibitive to sample from the target domain, or the reward signal can be sparse or delayed.
    %
    Examples include adapting an auto-driving agent pre-trained in simulated platforms to real environments~\cite{finn2019meta},
    %
    %
    The accessibility of information in the target domain can affect the way that TL approaches are designed.

    \item \noindent \textbf{\emph{How sample-efficient the TL approach is:}}
    %
    TL enables the RL  with better initial performance, hence usually requires fewer interactions compared with learning from scratch.
    Based on the sampling cost, we can categorize TL approaches into the following classes:
    (i) \emph{Zero-shot} transfer, which learns an agent that is directly applicable to the target domain without requiring any training interactions;
    (ii) \emph{Few-shot} transfer, which only requires a few samples (interactions) from the target domain;
    (iii) \emph{Sample-efficient} transfer, where an agent can benefit by TL to be more sample efficient compared to normal RL.
    %

     
\end{enumerate}

\subsection{Case Analysis of Transfer Learning in the context of Reinforcement Learning} \label{sec:difference} \vspace{-0.05in}
We now use \textit{HalfCheetah}\footnote{https://gym.openai.com/envs/HalfCheetah-v2/} as a working example to illustrate how TL can occur between the source and the target domain.
\textit{HalfCheetah} is a standard DRL benchmark for solving physical locomotion tasks, in which the objective is to train a two-leg agent to run fast without losing control of itself.

\vspace{-0.1in}
\subsubsection{Potential Domain Differences:}
During TL, the differences between the source and target domain may reside in any component of an MDP:
%
%

\begin{itemize}[leftmargin=*]
    \item $\cS$ (State-space): domains can be made different by extending or constraining the available positions for the \textit{HalfCheetah} agent to move.
    \item $\cA$ (Action-space) can be adjusted by changing the range of available torques for the thigh, shin, or foot of the agent.
    \item $\cR$ (Reward function): a domain can be simplified by using only the distance moved forward as rewards or be perplexed by using the scale of accelerated velocity in each direction as extra penalty costs.
    \item $\cT$ (Transition dynamics): two domains can differ by following different physical rules, leading to different transition probabilities given the same state-action pairs.
    \item $\mu_0$ (Initial states): the source and target domains may have different initial states, specifying where and with what posture the agent can start moving.
    \item $\tau$ (Trajectories): the source and target domains may allow a different number of steps for the agent to move before a task is done.
\end{itemize}

\subsubsection{Transferrable Knowledge:}
Without losing generality, we list below some transferrable knowledge assuming that the source and target domains are variants of \textit{HalfCheetah}:
%
\begin{itemize}[leftmargin=*]
    \item \textbf{\textit{Demonstrated trajectories}}: the target agent can learn from the behavior of a pre-trained expert, \eg~a sequence of running demonstrations.
    \item \textbf{\textit{Model dynamics}}: the RL agent may access a model of the physical dynamics for the source domain that is also partly applicable to the target domain. It can  perform dynamic programming based on the physical rules, running fast without losing its control due to the accelerated velocity.
    \item \textbf{\textit{Teacher policies}}: an expert policy may be consulted by the learning agent, which outputs the probability of taking different actions upon a given state example.
    \item \textbf{\textit{Teacher value functions}}: besides teacher policy, the learning agent may also refer to the value function derived by a teacher policy, which implies the quality of state-actions from the teacher's point of view. 
\end{itemize}

\vspace{-0.05in}
\subsection{Evaluation metrics} \label{sec:metrics} \vspace{-0.05in}
In this section, we present some representative metrics for evaluating TL approaches, which have also been partly summarized in prior work~\cite{taylor2009transfer,taylor2007transfer}:
\begin{itemize}[leftmargin=*]
    \item \emph{Jumpstart performance( jp)}: the initial performance (returns) of the agent.
    \item \emph{Asymptotic performance (ap)}: the ultimate performance (returns) of the agent.
    \item \emph{Accumulated rewards (ar)}: the area under the learning curve of the agent. 
    \item \emph{Transfer ratio (tr)}: the ratio between \textit{asymptotic performance} of the agent with TL and \textit{asymptotic performance} of the agent without TL.
    \item \emph{Time to threshold (tt)}: the learning time (iterations) needed for the target agent to reach certain performance threshold.
    \item \emph{Performance with fixed training epochs (pe)}: the performance achieved by the target agent after a specific number of training iterations.
    \item \emph{Performance sensitivity (ps)}: the variance in returns using different hyper-parameter settings. 
\end{itemize}

The above criteria mainly focus on the \textbf{\emph{learning process}} of the target agent.
In addition, we introduce the following metrics from the perspective of \textbf{\emph{transferred knowledge}}, which, although commensurately important for evaluation, have not been explicitly discussed by prior art:

\begin{itemize}[leftmargin=*]
    \item \emph{Necessary knowledge amount (nka)}: the necessary \emph{amount} of the knowledge required for TL in order to achieve certain performance thresholds. Examples along this line include the number of designed source tasks \cite{barreto2019transfer}, the number of expert policies, or the number of demonstrated interactions \cite{NEURIPS2020_92977ae4} required to enable knowledge transfer.
    \item \emph{Necessary knowledge quality (nkq)}: the guaranteed \emph{quality} of the knowledge required to enable effective TL. This metric helps in answering questions such as (i) Does the TL approach rely on near-oracle knowledge, such as expert demonstrations/policies \cite{ho2016generative}, or (ii) is the TL technique feasible even given suboptimal knowledge \cite{zhu2020learning}? 
%
\end{itemize}
%
TL approaches differ in various perspectives, including the forms of transferred knowledge, the RL frameworks utilized to enable such transfer, and the gaps between the source and the target domain. 
It maybe biased to evaluate TL from just one viewpoint. We believe that explicating these TL related metrics helps in designing more generalizable and efficient TL approaches.

In general, most of the abovementioned metrics can be considered as evaluating two  abilities of a TL approach: the \emph{\textbf{mastery}} and \emph{\textbf{generalization}}.
\emph{Mastery} refers to how well the learned agent can ultimately perform in the target domain, while \emph{generalization} refers to the ability of the learning agent to quickly adapt to the  target domain.
%

\judycom{
\section{Related Work} \label{sec:related_work} \vspace{-0.05in}
There are prior efforts in summarizing TL research in RL.
One of the earliest literatures is \cite{taylor2009transfer} .
Their main categorization is from the perspective of \emph{problem setting}, in which the TL scenarios may vary in the number of  domains involved, and the difference of state-action space among domains.
Similar categorization is adopted by \cite{lazaric2012transfer}, with more refined analysis dimensions including the objective of TL.
As pioneer surveys for TL in RL, neither \cite{taylor2009transfer} nor \cite{lazaric2012transfer} covered recent research  over the last decade.
For instance, \cite{taylor2009transfer} emphasized on different \emph{task-mapping} methods, which are more suitable for domains with tabular or mild state-action space dimensions.

There are other surveys  focused on specific subtopics that interplay between RL and TL.
For instance, \cite{zhao2020sim} consolidated sim-to-real TL methods. 
They explored work that is more tailored for \emph{robotics} domains, including domain generalization and zero-shot transfer, which is a favored application field of DRL as we discussed in Sec \ref{sec:applications}.
\cite{muller2021procedural} conducted extensive database search and summarized  benchmarks for evaluating TL algorithms in RL.
\cite{vithayathil2020survey} surveyed recent progress in multi-task RL. They partially shared research focus with us by studying certain TL oriented solutions towards multi-task RL, such as learning shared representations, pathNets, etc.
We surveyed TL for RL with a broader spectrum in methodologies, applications, evaluations, which naturally draws connections to the above literatures.
}

\begin{figure*}[htbp!]
  \centering{
      \includegraphics[width=0.7\linewidth]{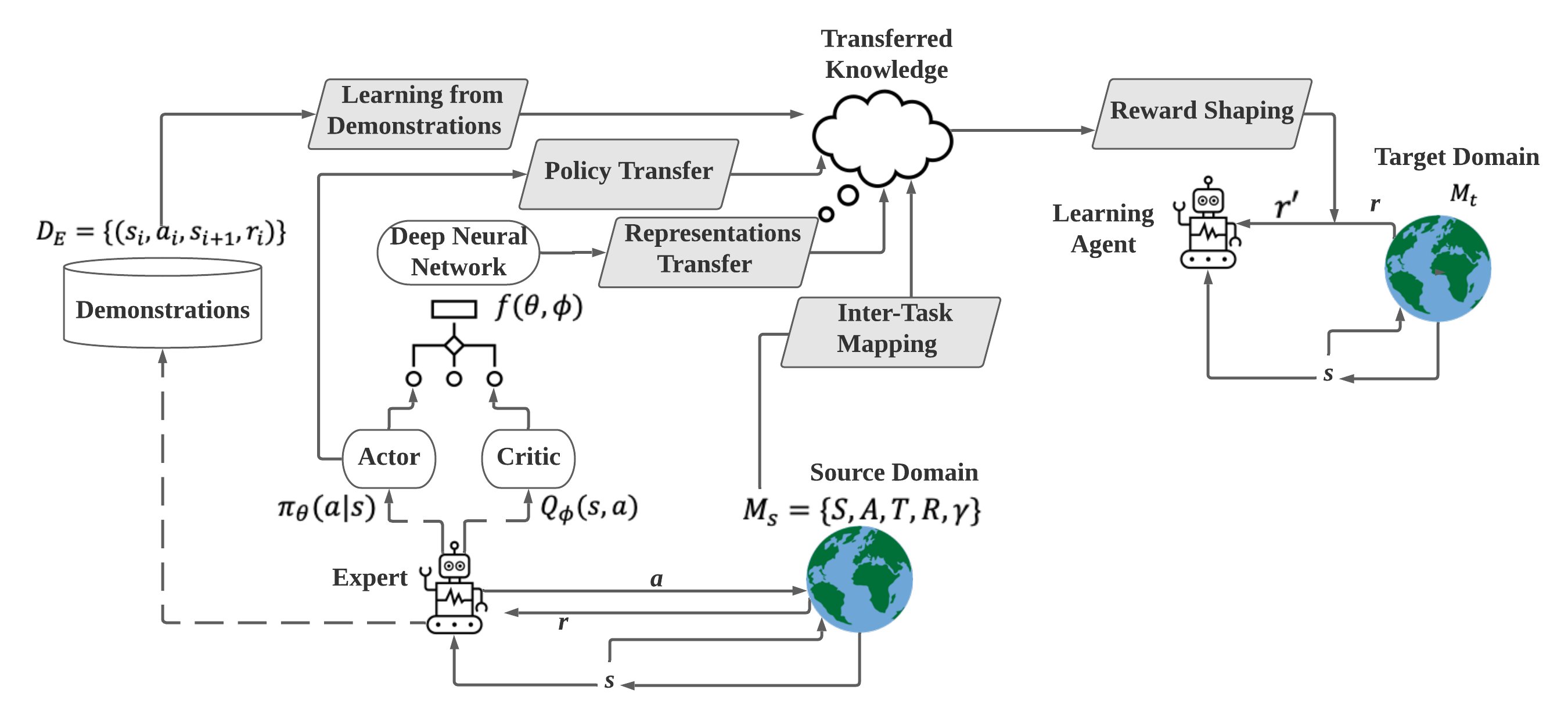}}
      \vspace{-0.1in}
      \caption{An overview of different TL approaches, organized by the format of transferred knowledge.}
      \label{fig:TL-framework} 
\end{figure*}
\vspace{-0.1in}

\section{Transfer Learning Approaches Deep Dive} \label{sec:approaches} \vspace{-0.05in}
In this section, we elaborate on various TL approaches and organize them into different sub-topics, mostly by answering the question of \textit{``what knowledge is transferred"}.
For each type of TL approach, we analyze them by following the other criteria mentioned in Section~\ref{sec:evaluation} and \judycom{and summarize the key evaluation metrics that are applicable to the discussed work.}
Figure~\ref{fig:TL-framework} presents an overview of different TL approaches discussed in this survey.

\subsection{Reward Shaping} \label{sec:rs} \vspace{-0.05in}
\modify{We start by introducing the \textit{Reward Shaping} approach, as it is applicable to most RL backbones and also largely overlaps with the other TL approaches discussed later.}
Reward Shaping (RS) is a technique that leverages the exterior knowledge to reconstruct the reward distribution of the target domain to guide the agent's policy learning.
More specifically, in addition to the environment reward signals, RS learns a reward-shaping function $\cF: \cS \times \cS \times \cA \to \RR $ to render auxiliary rewards, provided that the additional rewards contain external knowledge to guide the agent for better action selections.
Intuitively, an RS strategy will assign higher rewards to more beneficial state-actions to navigate the agent to desired trajectories.
As a result, the agent will learn its policy using the newly shaped rewards $\cR'$:
  $  \cR' = \cR + \cF $, which means that RS has altered the target domain with a different reward function:
\vspace{-0.05in}
\begin{align}
  \cM =(\cS, \cA, \cT, \gamma, \cR)) \to \cM' = (\cS, \cA, \cT, \gamma, \cR'). 
  \vspace{-0.1in}
\end{align}

Along the line of RS,
{\emph{Potential based Reward Shaping (PBRS)}} is one of the most classical approaches.
\cite{ng1999policy} proposed PBRS  to form a shaping function $F$ as the difference between two \textit{potential functions} ($\Phi(\cdot)$):
  \vspace{-.05in}
\begin{align} 
  F(s, a, s') = \gamma\Phi(s') - \Phi(s),
  \vspace{-.05in}
\end{align}
where the potential function $\Phi(\cdot)$ comes from the knowledge of expertise and evaluates the quality of a given state.
%
It has been proved that, without further restrictions on the underlying MDP or the shaping function $F$, PBRS is sufficient and necessary to preserve the policy invariance.
Moreover,  the optimal $Q$-function in the original and transformed MDP are related by the potential function:
$     Q_{\cM'}^*(s, a) = Q_{\cM}^*(s, a) - \Phi(s)$, 
which draws a connection between potential based reward-shaping and advantage-based learning approaches~\cite{williams1993tight}. 

The idea of \emph{PBRS} was extended to~\cite{wiewiora2003principled}, which formulated the potential as a function over both the state and the action spaces. This approach is called {\emph{Potential Based state-action Advice (PBA)}}.
The potential function $\Phi(s, a)$ therefore evaluates how beneficial an action $a$ is to take from state $s$:
\vspace{-.1in}
\begin{align} \label{eq:potential2}
    F(s, a, s', a') = \gamma \Phi(s', a') - \Phi(s, a).
\end{align}
\textit{PBA} requires on-policy learning and can be \textit{sample-costly}, as in Equation~(\ref{eq:potential2}), $a'$ is the action to take upon state $s$ is transitioning to $s'$ by following the learning policy. 
%
 
Traditional RS approaches assumed a \textit{static} potential function, until
\cite{devlin2012dynamic} proposed a {\emph{Dynamic Potential Based (DPB)}} approach which makes the potential a function of both states and time:
$
    F(s, t, s', t') = \gamma\Phi(s', t') - \Phi(s,t).
$
They proved that this dynamic approach can still maintain policy invariance:
$
    Q^*_{\cM'}(s,a) = Q^*_\cM(s, a) - \Phi(s, t), 
$
where $t$ is the current tilmestep. 
\cite{harutyunyan2015expressing} later introduced a way to incorporate any prior knowledge into a dynamic potential function structure, which is called \textbf{\emph{Dynamic Value Function Advice (DPBA)}}.
The rationale behind DPBA is that, given any extra reward function $R^+$ from prior knowledge, in order to add this extra reward to the original reward function,  the potential function should satisfy: 
  $\gamma \Phi(s', a') - \Phi(s, a) = F(s, a) = R^+(s,a).$

If $\Phi$ is not static but learned as an extra state-action \textit{Value} function overtime, then the Bellman equation for $\Phi$ is :
$     \Phi^{\pi}(s,a) = r^{\Phi}(s, a) + \gamma \Phi(s', a').$
The shaping rewards $F(s,a)$ is therefore the negation of $r^{\Phi}(s,a)$ :
\begin{align}
    F(s,a) = \gamma \Phi(s', a') - {\Phi}(s, a) = - r^{\Phi}(s,a).
\end{align}
This leads to the approach of using the negation of $R^+$ as the immediate reward to train an extra state-action \textit{Value} function $\Phi$ and the policy simultaneously.
Accordingly, the dynamic potential function $F$ becomes:
\begin{align}
    F_t(s,a) = \gamma \Phi_{t+1}(s', a') - {\Phi_t}(s, a).   
  \vspace{-.05in}
\end{align}
The advantage of \textit{DPBA} is that it provides a framework to allow arbitrary knowledge to be shaped as auxiliary rewards. 
 
Research along this line mainly focus on designing different shaping functions $F(s,a)$, while not much work has tackled the question of \textit{what knowledge can be used to derive this potential function}.
One work by~\cite{brys2015policy} proposed to use RS to transfer an expert policy from the source domain $\cM_s$ to the target domain $\cM_t$.
This approach assumed the existence of two mapping functions $M_S$ and $M_A$ that can transform the state and action from the source to the target domain.
%
%
Another work used demonstrated state-action samples from an expert policy to shape rewards~\cite{vevcerik2017leveraging}.
Learning the augmented reward involves learning a discriminator to distinguish samples generated by an expert policy from samples generated by the target policy. 
The loss of the discriminator is applied to shape rewards to incentivize the learning agent to mimic the expert behavior.
This work combines two TL approaches: RS and \textit{Learning from Demonstrations}, the latter of which will be elaborated in Section~\ref{sec:lfd}.

The above-mentioned RS approaches are summarized in Table~\ref{table:rs}.
%
They follow the potential based RS principle that has been developed systematically: 
from the classical \emph{PBRS} which is built on a \emph{static} potential shaping function of \emph{states},  
to \emph{PBA} which generates the potential as a function of both \emph{states} and \emph{actions}, 
and \emph{DPB} which learns a dynamic potential function of \emph{states} and \emph{time}, to the most recent \emph{DPBA}, which involves a dynamic potential function of \emph{states} and \emph{actions} to be learned as an extra state-action \textit{Value} function in parallel with the environment \textit{Value} function. 
As an effective TL paradigm, RS has been widely applied to fields including robot training \cite{tenorio2010robot}, spoken dialogue systems \cite{su2015reward}, and question answering \cite{lin2018multi}.  
It provides a feasible framework for transferring knowledge as the augmented reward and is generally applicable to various RL algorithms.
\modify{RS has also been applied to \textit{multi-agent} RL~\cite{devlin2014potential} and \textit{model-based} RL~\cite{grzes2009learning}.
Principled integration of RS with other TL approaches, such as \emph{Learning from demonstrations} (Section \ref{sec:lfd}) and \emph{Policy Transfer} (Section \ref{sec:policyTransfer}) will be an intriguing question for ongoing research.

Note that RS approaches discussed so far are built upon a consensus that the source information for shaping the reward comes \emph{externally}, which coincides with the notion of \textit{knowledge transfer}.
Some RS work also tackles the scenario where the shaped reward comes \emph{intrinsically}. 
For instance, \emph{Belief Reward Shaping} was proposed by \cite{marom2018belief}, which utilizes a Bayesian reward shaping framework to generate the potential value that decays with experience, where the potential value comes from the critic itself.
%
%
}


%
\begin{table*}[htbp!]
    \centering
    \begin{tabular}{ccccc} \hline
      {\textbf{Methods}} &  
      {\textbf{MDP difference}} & 
      {\textbf{Format of shaping reward}} & 
      {\textbf{
        Knowledge source}} & 
      {\textbf{Evaluation metrics}}  
        \\ \hline   
       \textit{PBRS} 
     & { $\cM_s = \cM_t$ } 
     & { $F = \gamma \Phi(s') - \Phi(s)$ } 
     & { \xmark } 
     & { \emph{ap, ar} } 
     \\ \hline 
       \textit{PBA} 
     & { $\cM_s = \cM_t$ } 
     & { $F = \gamma \Phi(s', a') - \Phi(s, a)$ } 
     & { \xmark } 
     & { \emph{ap, ar} } 
     \\ \hline
       \textit{DPB} 
     & { $\cM_s = \cM_t$ } 
     & { $F = \gamma\Phi(s', t') - \Phi(s,t)$ } 
     & { \xmark } 
     & { \emph{ap, ar} } 
     \\ \hline
     \textit{DPBA}
     & { $\cM_s = \cM_t$ } 
     & \multicolumn{1}{p{4cm}}{ $F_t = \gamma \Phi_{t+1}(s', a') - {\Phi_t}(s, a)$ , $\Phi$ learned as an extra Q function} 
     & { \xmark } 
     & { \emph{ap, ar} } 
     \\ \hline
       {\cite{brys2015policy}} 
     & { $\cS_s \neq \cS_t$, $\cA_s \neq \cA_t$  } 
     & { $F_t = \gamma \Phi_{t+1}(s', a') - {\Phi_t}(s, a)$ } 
     & { $\pi_s$} 
     & { \emph{ap, ar} } 
     \\ \hline 
       {\cite{vevcerik2017leveraging}} 
     & { $\cM_s = \cM_t$  } 
     & { $F_t = \gamma \Phi_{t+1}(s', a') - {\Phi_t}(s, a)$ } 
     & { $D_E$} 
     & { \emph{ap, ar} } 
     \\ \hline  
    \end{tabular}
    \vspace{-0.05in}
    \caption{A comparison of \textit{reward shaping} approaches. \xmark~denotes that the information is not revealed in the paper.\label{table:rs}}
    \label{tb:rsd}
    \end{table*} 
    \vspace{-0.1in} 

\vspace{-0.1in}
\subsection{Learning from Demonstrations} \label{sec:lfd}
%
\emph{Learning from Demonstrations (LfD)} is a technique to assist RL by utilizing external demonstrations for more efficient exploration.
%
%
The demonstrations may come from different sources with varying qualities.
%
%
Research along this line usually address a scenario where the source and the target MDPs are the same: $\cM_s = \cM_t$, although there has been work that learns from demonstrations generated in a different domain~\citep{liu2019state,kim2020domain}. 

Depending on \textbf{\emph{when}} the demonstrations are used for knowledge transfer, approaches can be organized into \emph{offline}  and \emph{online} methods.
For \emph{offline} approaches, demonstrations are either used for pre-training RL components, or for \textit{offline} RL \cite{ma2019imitation,yang2021representation}.
When leveraging demonstrations for pre-training, RL components such as the value function $V(s)$~\cite{zhang2018pretraining}, the policy $\pi$~\cite{silver2016mastering}, or  the model of transition dynamics~\cite{schaal1997learning}, can be initialized by learning from  demonstrations.
%
%
For the \emph{online} approach, demonstrations are directly used to guide agent actions for efficient explorations~\cite{hester2018deep}.
Most work discussed in this section follows the online transfer paradigm or combines offline pre-training with online RL~\cite{nair2018overcoming}.

Work along this line can also be categorized depending on \emph{\textbf{what}} RL frameworks are compatible:
some adopts the policy-iteration framework~\cite{chemali2015direct,kim2013learning,piot2014boosted}, 
some follow a $Q$-learning framework~\cite{hester2018deep,brys2015reinforcement}, while recent work usually follows the policy-gradient framework~\cite{vevcerik2017leveraging,nair2018overcoming,kang2018policy,zhu2020learning}.
Demonstrations have been leveraged in the {\emph{policy iterations}} framework by~\cite{bertsekas2011approximate}.
Later, \cite{chemali2015direct} introduced the {\emph{Direct Policy Iteration with Demonstrations (DPID)}} algorithm. 
This approach samples complete demonstrated rollouts $D_E$ from an expert policy $\pi_E$, in combination with the self-generated rollouts $D_{\pi}$ gathered from the learning agent.
$D_\pi \cup D_E$ are used to learn a Monte-Carlo estimation of the Q-value: $\hat{Q}$, from which a learning policy can be derived greedily: $\pi(s)= \underset{ a \in \cA} {\arg \max}\hat{Q}(s,a)$.
This policy $\pi$ is further regularized by a loss function $\cL(s,\pi_E)$ to minimize its discrepancy from the expert policy decision.

Another example is the {\emph{Approximate Policy Iteration with Demonstration (APID)}} algorithm, which was proposed by~\cite{kim2013learning} and extended by~\cite{piot2014boosted}.
Different from \emph{DPID} where both $D_E$ and $D_\pi$ are used for value estimation, the \emph{APID} algorithm solely applies $D_\pi$ to approximate on the Q function.
Expert demonstrations $D_E$ are used to learn the value function, which, given any state $s_i$, renders expert actions $\pi_E(s_i)$ with higher $Q$-value margins compared with other actions that are not shown in $D_E$: 
\begin{align}
    Q(s_i, \pi_E(s_i)) - \underset{a \in \cA \backslash \pi_E(s_i)}{\max}Q(s_i, a) \geq 1 - \xi_i. 
\vspace{-0.05in}
\end{align}
The term $\xi_i $ is used to account for the case of imperfect demonstrations. 
\cite{piot2014boosted} further extended the work of \emph{APID} with a different evaluation loss:
\begin{align}
  \cL^\pi = \cE_{(s,a) \sim {D_{\pi}}} \Vert\cT^* Q(s,a)-Q(s,a) \Vert,
\end{align}
where $\cT^* Q(s,a) = R(s,a) + \gamma \cE_{s'\sim p(.|s,a)} [ \underset{a'}{\max} Q(s', a')].$
Their work theoretically converges to the optimal $Q$-function compared with \emph{APID}, as $\cL_{\pi}$ is minimizing the optimal Bellman residual instead of the empirical norm.

In addition to policy iteration, the following two approaches integrate demonstration data into the TD-learning framework, such as {{$Q$-learning}}.
Specifically, \cite{hester2018deep} proposed the {\emph{DQfD}} algorithm, which maintains two separate replay buffers to store demonstrated data and self-generated data, respectively, so that expert demonstrations can always be sampled with a certain probability.
Their method leverages the refined priority replay mechanism~\cite{schaul2016prioritize} where the probability of sampling a transition $i$ is based on its priority $p_i$ with a temperature parameter $\alpha$:
$P(i) = \frac{p_i^{\alpha}}{\sum_{k}{p_k^\alpha}}.$ 
\modify{
Another algorithm named {\textit{LfDS}}  was proposed by~\cite{brys2015reinforcement}, which draws a close connection to  \textit{reward shaping} (Section \ref{sec:rs}).
%
\textit{LfDS} builds the potential value of a  \textit{state-action} pair as the highest similarity between the given pair and the expert demonstrations.}
This augmented reward assigns more credits to state-actions that are more similar to expert demonstrations, encouraging the agent for expert-like behavior.

Besides $Q$-learning, recent work has integrated \textit{LfD} into \textbf{\emph{policy gradient}}~\cite{vevcerik2017leveraging,nair2018overcoming,ho2016generative,kang2018policy,zhu2020learning}.
A representative work along this line is {\textit{Generative Adversarial Imitation Learning (GAIL)}}~\cite{ho2016generative}. 
\textit{GAIL} introduced the notion of \emph{occupancy measure} $d_\pi$, which is the stationary state-action distributions derived from a policy $\pi$.
Based on this notion, a new reward function is designed such that maximizing the accumulated new rewards encourages minimizing the distribution divergence between the \emph{occupancy measure} of the current policy $\pi$ and the expert policy $\pi_E$. 
Specifically, the new reward is learned by adversarial training~\cite{goodfellow2014generative}: a discriminator $D$ is learned to  distinguish interactions sampled from the current policy $\pi$ and the expert policy $\pi_E$:
\begin{small}
\begin{align}
  J_D = \max_{D:\cS \times \cA \to (0,1)} \cE_{d_\pi}\log[1-D(s,a)] + \cE_{d_E}\log[D(s,a)]
\end{align}
\end{small}
Since $\pi_E$ is unknown, its state-action distribution $d_E$ is estimated based on the given expert demonstrations $D_E$.
%
%
The output of the discriminator is used as new rewards to encourage distribution matching, with $r'(s,a) = -\log(1-D(s,a))$.
The RL process is naturally altered to perform distribution matching by min-max optimization:
\begin{align*}
  \max_\pi \min_D J(\pi, D): &=   \cE_{d_\pi}\log[1-D(s,a)] + \cE_{d_E}\log[D(s,a)].
\end{align*} 

The philosophy in \emph{GAIL} of using expert demonstrations for distribution matching has inspired other \textit{LfD} algorithms.
For example, \cite{kang2018policy} extended  \textit{GAIL} with an algorithm called  \emph{Policy Optimization from Demonstrations (POfD)}, which combines the discriminator reward with the environment reward:
\begin{align} \label{objective:pofd} 
  \max_\theta  = \cE_{d_\pi}[r(s,a)] - \lambda D_{JS}[d_\pi || d_E].
\end{align}

%

Both \textit{GAIL} and \textit{POfD} are under an \emph{on-policy} RL framework.
To further improve the sample efficiency of TL, some \emph{off-policy} algorithms have been proposed, such as \emph{DDPGfD}~\cite{vevcerik2017leveraging} which is built upon the \textit{DDPG} framework.
\textit{DDPGfD} shares a similar idea as \emph{DQfD} in that they both use a second replay buffer for storing demonstrated data, and each demonstrated sample holds a sampling priority $p_i$.
For a demonstrated sample, its priority $p_i$ is augmented with a constant bias $\epsilon_D > 0$ for encouraging more frequent sampling of expert demonstrations:
\begin{align*} 
    p_i = \delta_i^2 + \lambda\|\nabla_a Q(s_i, a_i| \theta^Q)\|^2 + \epsilon + \epsilon_D,
\end{align*}
where $\delta_i$ is the TD-residual for transition, $\|\nabla_a Q(s_i, a_i| \theta^Q)\|^2$ is the loss applied to the actor,  and $\epsilon$ is a small
positive constant to ensure all transitions are sampled with some probability.
Another work also adopted the \textit{DDPG} framework to learn from demonstrations \cite{nair2018overcoming}. 
Their approach differs from \textit{DDPGfD} in that its objective function is augmented with a \emph{Behavior Cloning Loss} to encourage imitating on provided demonstrations:
    $\cL_{BC} = \sum_{i=1}^{|D_E|}||\pi(s_i | \theta_\pi) - a_i||^2$.

%
To further address the issue of \textit{suboptimal} demonstrations, in \cite{nair2018overcoming} the form of \emph{Behavior Cloning Loss} is altered based on the critic output, so that only demonstration actions with higher $Q$ values will lead to the loss penalty:
\begin{small}
\begin{align}
    \cL_{BC} = \sum_{i=1}^{|D_E|}\left\|\pi(s_i | \theta_\pi) - a_i\right\|^2 \cone[Q(s_i, a_i) > Q(s_i, \pi(s_i))].
\end{align}
\end{small}
%

There are several challenges faced by \textit{LfD}, one of which is the \emph{imperfect demonstrations}.
Previous approaches usually presume near-oracle demonstrations.
%
%
Towards tackling  \textit{suboptimal} demonstrations,
\cite{kim2013learning} leveraged the hinge-loss function to allow occasional violations of the property that $Q(s_i, \pi_E(s_i)) - \underset{a \in \cA \backslash \pi_E(s_i)}{\max}Q(s_i, a) \geq 1$. 
Some other work uses regularized objective to alleviate overfitting on biased data~\cite{hester2018deep,schaul2016prioritize}. 
A different strategy is to leverage those sub-optimal demonstrations only to boost the initial learning stage.
For instance, \cite{zhu2020learning} proposed \emph{Self-Adaptive Imitation Learning (SAIL)}, which learns from suboptimal demonstrations using generative adversarial training while gradually selecting self-generated trajectories with high qualities to replace less superior demonstrations. 

Another challenge faced by \textit{LfD} is \emph{covariate drift} (\cite{ross2011reduction}): demonstrations may be provided in limited numbers, which results in the learning agent lacking guidance on states that are unseen in the demonstration dataset. 
This challenge is aggravated in MDPs with sparse reward feedbacks, as the learning agent cannot obtain much supervision information from the environment either. 
%
Current efforts to address this challenge include encouraging explorations by using an entropy-regularized objective~\cite{gao2018reinforcement}, decaying the effects of demonstration guidance by softening its regularization on policy learning over time~\cite{jing2020reinforcement}, and introducing \emph{disagreement regularizations} by training an ensemble of policies based on the given demonstrations, where the variance among policies serves as a  negative reward function \cite{brantley2019disagreement}. 
 
We summarize the above-discussed approaches in Table~\ref{table:lfd}. 
In general, demonstration data can help in both \emph{offline} pre-training for better initialization and \emph{online} RL for efficient exploration.
During the RL phase, demonstration data can be used together with self-generated data to encourage expert-like behaviors (\emph{DDPGfD, DQFD}), to shape value functions (\emph{APID}), or to guide the policy update in the form of an auxiliary objective function (\emph{PID,GAIL, POfD}).
%
%
%
\judycom{
To validate the algorithm robustness given different knowledge resources, most \emph{LfD} methods are evaluated using metrics that either indicate the performance under \emph{limited} demonstrations (\emph{nka}) or \emph{suboptimal} demonstrations (\emph{nka}). 
The integration of \emph{LfD} with \emph{off-policy} RL backbone makes it natural to adopt \emph{pe} metrics for evaluating how learning efficiency can be further improved by knowledge transfer.
}
Developing more general \textit{LfD} approaches that are agnostic to RL frameworks and can learn from sub-optimal or limited demonstrations would be the ongoing focus for this research domain.

\begin{table*}[htbp!]
\centering
\begin{tabular}{cccccc} \hline
  \multicolumn{1}{c}{\textbf{Methods}} &  \multicolumn{1}{c} {\begin{tabular}[c]{@{}c@{}} \textbf{Optimality} \\ \textbf{guarantee} \end{tabular}}  & \multicolumn{1}{l}{\textbf{Format of transferred demonstrations}} & \multicolumn{1}{c}{\textbf{RL framework}} & \multicolumn{1}{c}{\judycom{\textbf{Evaluation metrics}}} \\ \toprule

   \textit{DPID} 
 & { \cmark} 
 & \multicolumn{1}{p{6cm}}{ Indicator binary-loss : $\cL(s_i)=\cone\{ \pi_E(s_i) \neq \pi(s_i)\}$ }  
 & \textit{API }
 & {\judycom{ \emph{ap, ar, nka} }}
 \\ \hline
   \textit{APID} 
 & { \xmark} 
 & \multicolumn{1}{p{6cm}}{ Hinge loss on the marginal-loss: $\big [ \cL(Q, \pi, \pi_E)\big ]_{+}$ } 
 & \textit{API}
 & {\judycom{   \emph{ap, ar, nta, nkq} } }
 \\ \hline
   \textit{APID} extend 
 & { \cmark} 
 & \multicolumn{1}{p{8cm}}{Marginal-loss: $\cL(Q, \pi, \pi_E)$} 
 & \textit{API}
 & {\judycom{   \emph{ap, ar, nta, nkq} } }
 \\ \hline
 
 { \cite{nair2018overcoming}} 
 & { \cmark} 
 & \multicolumn{1}{p{8cm}}{ Increasing sampling priority and behavior cloning loss} 
 & \textit{DDPG }
 & {\judycom{  \emph{ap, ar, tr, pe, nkq} }}
 \\ \hline
   { \textit{DQfD}} 
 & { \xmark } 
 & \multicolumn{1}{p{6cm}}{Cached transitions in the replay buffer } 
 & {  \textit{DQN} } 
 & {\judycom{   \emph{ap, ar, tr}}}
 \\ \hline
    \textit{LfDS} 
 & { \xmark } 
 & \multicolumn{1}{p{6cm}}{ Reward shaping function } 
 & {  \textit{DQN} }
 & {\judycom{   \emph{ap, ar, tr} } }
\\ \hline
 \textit{GAIL} 
 & { \cmark } 
 & \multicolumn{1}{p{6cm}}{ Reward shaping function: $-\lambda \log(1 - D(s, a))$ } 
 & \textit{TRPO}
 &{ \judycom{   \emph{ap, ar, tr, pe, nka} } }
 \\ \hline
   \textit{POfD} 
   & { \cmark } 
 & \multicolumn{1}{p{4cm}}{ Reward shaping function: $r(s,a)-\lambda \log(1 - D(s, a))$ } 
 & \textit{TRPO,PPO}
 & {\judycom{   \emph{ap, ar, tr, pe, nka} } }
 \\ \hline
   \textit{DDPGfD} 
(pe) & { \cmark} 
 & \multicolumn{1}{p{6cm}}{ Increasing sampling priority} 
 & \textit{DDPG}
 & {\judycom{   \emph{ap, ar, tr, pe} } }
 \\ \hline

 \textit{SAIL} 
 & { \xmark} 
 & \multicolumn{1}{p{8cm}}{Reward shaping function: $r(s,a)-\lambda \log(1 - D(s, a))$ }   
 & \textit{DDPG}
 & {\judycom{   \emph{ap, ar, tr, pe, nkq, nka} } }
 \\ \hline
 
\end{tabular}
\vspace{-0.1in}
\caption{A comparison of \textit{learning from demonstration} approaches.\label{table:lfd}}
\label{tb:lfd}
\end{table*}

\vspace{-0.1in}
\subsection{Policy Transfer} \label{sec:policyTransfer} \vspace{-0.05in}
\textit{Policy transfer} is a TL approach where the external knowledge takes the form of pre-trained policies from one or multiple source domains.
Work discussed in this section is built upon a \emph{many-to-one} problem setting, described as below:
\begin{policyTransfer*}
    A set of teacher policies $\pi_{E_1}, \pi_{E_2}, \dots, \pi_{E_K}$ are trained on a set of source domains $\cM_1, \cM_2, \dots, \cM_K$, respectively. 
    A student policy $\pi$ is learned for a target domain by leveraging knowledge from $\{\pi_{E_i}\}_{i=1}^K$.
\end{policyTransfer*}
For the \emph{one-to-one} scenario with only one teacher policy, one can consider it as a special case of the above with $K=1$. Next, we categorize recent work of policy transfer into two techniques: \emph{policy distillation} and \emph{policy reuse}.

\vspace{-0.05in}
\subsubsection {Transfer Learning via Policy Distillation}
\modify{The idea of \emph{knowledge distillation} has been applied to the field of RL to enable \textit{policy distillation}.
%
 {Knowledge distillation} was first proposed by~\cite{hinton2014distilling} as an approach of knowledge ensemble from multiple teacher models into a single student model.
}
%
%
Conventional policy distillation approaches transfer the teacher policy following a supervised learning paradigm~\cite{rusu2015policy, yin2017knowledge}.
Specifically, a student policy is learned by minimizing the divergence of action distributions between the teacher policy $\pi_E$ and student policy $\pi_{\theta}$, which is denoted as $\cH^{\times}(\pi_E(\tau_t) | \pi_\theta(\tau_t))$:   
\begin{align}
    \min_{\theta}\cE_{\tau \sim \pi_E}[\sum_{t=1}^{|\tau|} \nabla_{\theta} \cH^{\times}(\pi_E(\tau_t) | \pi_\theta(\tau_t))].
\end{align}
The above expectation is taken over trajectories sampled from the teacher policy $\pi_E$, hence this approach is called \emph{teacher distillation}. 
One example along this line is~\cite{rusu2015policy}, in which $N$ teacher policies are learned for $N$ source tasks separately, 
and each teacher yields a dataset $D^E=\{s_i, \vq_i\}_{i=0}^N$ consisting of observations $s$ and vectors of the  corresponding $Q$-values $\vq$, such that $\vq_i=[Q(s_i,a_1),Q(s_i,a_2),...|a_j \in \cA]$.  
Teacher policies are further distilled to a single student $\pi_\theta$ by minimizing the KL-Divergence between each teacher $\pi_{E_i}(a|s)$ and the student $\pi_\theta$, approximated using the dataset $D^E$:
$\min_{\theta}\cD_{KL}(\pi^E | \pi_\theta) \approx \sum_{i=1}^{|D^E|} \text{softmax}\left(\frac{\vq^E_i}{\tau}\right) \ln \left(\frac{\text{softmax}(\vq_i^E)}{\text{softmax}(\vq_i^\theta)}\right)$.

Another policy distillation approach is \emph{student distillation}~\cite{czarnecki2019distilling,parisotto2015actor}, which is resemblant to teacher distillation except that during the optimization step, the objective expectation is taken over trajectories sampled from the student policy instead of the teacher policy, \ie:
$\min_{\theta}\cE_{\tau \sim \pi_\theta}\left[\sum_{t=1}^{|\tau|} \nabla_{\theta} \cH^{\times}(\pi_E(\tau_t) | \pi_\theta(\tau_t))\right]$.
\cite{czarnecki2019distilling} summarized related work on both kinds of distillation approaches.
Although it is feasible to combine both distillation approaches~\cite{ross2011reduction},
we observe that more recent work focuses on student distillation, which empirically shows better exploration ability compared to teacher distillation, especially when the teacher policies are \emph{deterministic}.

Taking an alternative perspective, there are two approaches of policy distillation:
(1) minimizing the cross-entropy between the teacher and student policy distributions over actions \cite{parisotto2015actor, schmitt2018kickstarting};
and (2) maximizing the probability that the teacher policy will visit trajectories generated by the student,  \ie\ $\max_{\theta} P(\tau \sim \pi_E | \tau \sim \pi_\theta)$ \cite{teh2017distral, schulman2017equivalence}.
One example of approach (1) is the \emph{Actor-mimic} algorithm \cite{parisotto2015actor}.
This algorithm distills the knowledge of expert agents into the student by minimizing the cross entropy between the student policy $\pi_\theta$ and each teacher policy $\pi_{E_i}$ over actions: 
    $\cL^i(\theta) = \sum_{a \in \cA_{E_i}} \pi_{E_i}(a|s) \log_{\pi_\theta}(a|s)$, 
where each teacher agent is learned using a \textit{DQN} framework. The teacher policy is therefore derived from the Boltzmann distributions over the $Q$-function output:
%
    $\pi_{E_i}(a|s) = \frac{e^{\tau^{-1} Q_{E_i}(s,a)}}{\sum_{a' \in \cA_{E_i}} e^{\tau^{-1} Q_{E_i}(s,a')}}$.
%
An instantiation of approach (2) is the \emph{Distral} algorithm \cite{teh2017distral}.
which learns a \emph{centroid} policy $\pi_\theta$ that is derived from  $K$ teacher policies.
%
The knowledge in each teacher $\pi_{E_i}$ is distilled to the centroid and get transferred to the student, while both the transition dynamics $\cT_i$ and reward distributions $\cR_i$ for source domain $\cM_i$ are heterogeneous.
The student policy is learned by maximizing a multi-task learning objective $\max_\theta \sum_{i=1}^K J(\pi_\theta, \pi_{E_i})$, where 
\begin{small}
\begin{align*}
    J(\pi_\theta, \pi_{E_i}) =  \sum_{t} &\cE_{(s_t,a_t)\sim \pi_\theta} \Big[ \sum_{t\geq 0} \gamma^t (r_i(a_t, s_t) +  \\
    &\frac{\alpha}{\beta} \log \pi_\theta(a_t|s_t) - \frac{1}{\beta} \log(\pi_{E_i}(a_t|s_t)) ) \Big], 
\end{align*}
\end{small}
in which both $\log \pi_\theta(a_t|s_t)$ and $\pi_\theta$ are used as augmented rewards. Therefore, the above approach also draws a close connection to \textit{Reward Shaping}  (Section \ref{sec:rs}). 
In effect, the $\log \pi_\theta(a_t|s_t)$ term guides the learning policy $\pi_\theta$ to yield actions that are more likely to be generated by the teacher policy,
whereas the entropy term $-\log(\pi_{E_i}(a_t|s_t)$ encourages exploration.
A similar approach was proposed by \cite{schmitt2018kickstarting} which only uses the cross-entropy between teacher and student policy $ \lambda \cH(\pi_{E}(a_t|s_t)|| \pi_\theta(a_t|s_t))$ to reshape rewards.
Moreover, they adopted a dynamically fading coefficient to alleviate the effect of the augmented reward so that the student policy becomes independent of the teachers after certain optimization iterations.

\subsubsection{Transfer Learning via Policy Reuse} \label{sec:GPI}
\modify{\emph{Policy reuse} directly reuses policies from source tasks to build the target policy.}
The notion of policy reuse was proposed by \cite{fernandez2006probabilistic}, which directly learns the target policy as a weighted combination of different source-domain policies, and the probability for each source domain policy to be used is related to its expected performance gain in the target domain:
$P(\pi_{E_i}) = \frac{\exp{(t W_i)}}{\sum_{j=0}^{K}\exp{(t W_j)}},$
where $t$ is a dynamic temperature parameter that increases over time. 
Under a $Q$-learning framework, the $Q$-function of the target policy is learned in an iterative scheme: 
during every learning episode, $W_i$ is evaluated for each expert policy $\pi_{E_i}$, and $W_0$ is obtained for the learning policy, from which a reuse probability $P$ is derived. 
Next, a behavior policy is sampled from this probability $P$. 
%
%
After each training episode, both $W_i$ and the temperature $t$ for calculating the reuse probability is updated accordingly.
One limitation of this approach is that the $W_i$, \ie~the expected return of each expert policy on the target task, needs to be evaluated frequently. This work was implemented in a tabular case, leaving the scalability issue unresolved.
More recent work by \cite{barreto2017successor} extended the \emph{policy improvement} theorem \cite{bellman1966dynamic} from one to multiple policies, which is named as \emph{Generalized Policy Improvement}.
We refer its main theorem as follows:  
\vspace{-0.05in}
\begin{theorem*}{\emph{[Generalized Policy Improvement (GPI)]}} 
Let $\{\pi_i\}_{i=1}^n$ be $n$ policies and let $\{\hat{Q}^{\pi_i}\}_{i=1}^n$ be their approximated action-value functions, s.t:
    $\Big | Q^{\pi_i}(s,a) - \hat{Q}^{\pi_i}(s,a) \Big | \leq \epsilon ~ \forall s \in \cS, a \in \cA \text{, and } i \in [n]$.    
Define $\pi(s) = \underset{a}{\arg \max} ~\underset{i}{\max}\hat{Q}^{\pi_i}(s,a)$, then: 
$ Q^\pi(s,a) \geq \underset{i}{\max}Q^{\pi_i}(s,a) - \frac{2}{1 - \gamma } \epsilon$, $\forall~s \in \cS, a \in \cA$.
\end{theorem*}

Based on this theorem, a policy improvement approach can be naturally derived by greedily choosing the action which renders the highest $Q$-value among all policies for a given state.
Another work along this line is \cite{barreto2017successor}, in which an expert policy $\pi_{E_i}$ is also trained on a different source domain $\cM_i$ with reward function $\cR_i$, so that $Q^{\pi}_{\cM_0}(s,a) \neq Q^{\pi}_{\cM_i}(s,a) $.
To efficiently evaluate the $Q$-functions of different source policies in the target MDP, a disentangled representation $\vpsi(s,a)$ over the states and actions is learned using neural networks and is generalized across multiple tasks. Next, a task (reward) mapper $\vW_i$ is learned, based on which the $Q$-function can be derived:
$Q^\pi_i(s,a) = \vpsi(s,a)^T \vW_i.$
\cite{barreto2017successor} proved that the loss of \textit{GPI} is bounded by the difference between the source and the target tasks.
In addition to policy-reuse, their approach involves learning a shared representation $\vpsi(s,a)$, which is also a form of transferred knowledge and will be elaborated more in Section \ref{sec:learnDisentangle}.  

We summarize the abovementioned policy transfer approaches in Table~\ref{table:policy-transfer}.
In general, policy transfer can be realized by \textit{knowledge distillation}, which can be either optimized from the student's perspecive (\textit{student distillation}), or from the teacher's perspective (\textit{teacher distillation})
Alternatively, teacher policies can also be directly \textit{reused} to update the target policy.
\judycom{
  Regarding evaluation, most of the abovementioned work has investigated a multi-teacher transfer scenario, hence the \emph{generalization} ability or \emph{robustness} is largely evaluated on metrics such as \emph{performance sensitivity(ps)} (\eg~ performance given different numbers of teacher policies or source tasks ). \emph{Performance with fixed epochs (pe)} is another commonly shared metric to evaluate how the learned policy can quickly adapt to the target domain.
}
All approaches discussed so far presumed one or multiple \textit{expert} policies, which are always at the disposal of the learning agent.  
Open questions along this line include \textit{How to leverage imperfect policies for knowledge transfer}, or \textit{How to refer to teacher policies within a budget}.

\begin{table*}[htbp!]
    \centering
    \begin{tabular}{ccccc} 
      \multicolumn{1}{c}{\textbf{Paper}} &  \multicolumn{1}{c}{\textbf{Transfer approach}} &  \multicolumn{1}{c}{\textbf{MDP difference}} & \multicolumn{1}{c}{\textbf{RL framework}}  & \multicolumn{1}{c}{\judycom{\textbf{Evaluation metrics}}} \\ \hline 
       { \cite{rusu2015policy}} 
     & { Distillation }  
     & {  $\cS, \cA$}
     & {  DQN}
     & { \judycom{ $ap, ar$}} \\ \hline 
     { \cite{yin2017knowledge}} 
     & { Distillation }  
     & {  $\cS, \cA$}
     & { DQN}
     & { \judycom{ $ap, ar, pe, ps$}}\\ \hline      
       { \cite{parisotto2015actor}} 
     & { Distillation }  
     & {  $\cS, \cA$}
     & { Soft Q-learning}
     & { \judycom{ $ap, ar, tr$, $pe, ps$}}\\ \hline
     { \cite{teh2017distral}} 
     & { Distillation }  
     & {  $\cS, \cA$}
     & { A3C}
     & { \judycom{ $ap, ar, pe$, $tt$}}\\ \hline      
       { \cite{fernandez2006probabilistic}} 
     & { Reuse }  
     & {  $\cR$}
     & { Tabular Q-learning}
     & { \judycom{ $ap, ar, ps, tr$}}\\ \hline   
       { \cite{barreto2017successor}} 
     & { Reuse}  
     & {  $\cR$}
     & { DQN}
     & { \judycom{ $ap, ar, pe, ps$}}\\ \hline   
    \end{tabular}
    \vspace{-0.05in}
    \caption{A comparison of \textit{policy transfer} approaches. \label{table:policy-transfer}}
    \label{tb:lfd}
    \vspace{-0.1in}
    \end{table*}  
    

\vspace{-0.05in}
\subsection{Inter-Task Mapping} \label{sec:mapping}
In this section, we review TL approaches that utilize \textit{\textbf{mapping functions}} between the source and the target domains to assist knowledge transfer.
Research in this domain can be analyzed from two perspectives:
(1) \emph{which domain does the mapping function apply to}, and (2) \emph{how is the mapped representation utilized}.
Most work discussed in this section shares a common assumption as below: 
\vspace{-0.05in}
\begin{assumption*} 
  One-to-one mappings exist between the source domain $\cM_s$ and the target domain $\cM_t$.
\end{assumption*}
%

Earlier work along this line requires a \emph{given mapping function}~\cite{taylor2007transfer,torrey2005using}.
One examples is~\cite{taylor2007transfer} which assumes that each target state (action) has a unique correspondence in the source domain, and two mapping functions $X_S, X_A$ are provided over the state space and the action space, respectively, so that $X_S(\cS^t) \to \cS^s$, $X_A(\cA^t) \to \cA^s$. Based on $X_S$ and $X_A$, a mapping function over the $Q$-values $M(Q_s) \to Q_t$ can be derived accordingly.
Another work is done by~\cite{torrey2005using} which transfers \emph{advice} as the knowledge between two domains.
In their settings, the \emph{advice} comes from a human expert who provides the mapping function over the $Q$-values in the source domain and transfers it to the learning policy for the target domain.
This {advice} encourages the learning agent to prefer certain good actions over others, which equivalently provides a relative ranking of actions in the new task.

More later research tackles the inter-task mapping problem by \emph{learning}  a mapping function~\cite{gupta2017learning,konidaris2006autonomous,ammar2012reinforcement}.
Most work learns a mapping function over the \textbf{\textit{state}} space or a subset of the state space.
In their work, state representations are usually divided into \emph{agent-specific} and \emph{task-specific} representations, denoted as $s_{agent}$ and $s_{env}$, respectively.
In~\cite{gupta2017learning} and~\cite{konidaris2006autonomous}, the mapping function is learned on the \emph{agent-specific} sub state, and the mapped representation is applied to reshape the immediate reward.
For~\cite{gupta2017learning}, the invariant feature space mapped from $s_{agent}$ can be applied across agents who have distinct action space but share some morphological similarity.
Specifically, they assume that both agents have been trained on the same \emph{proxy} task, based on which the mapping function is learned.
The mapping function is learned using an encoder-decoder  structure~\cite{badrinarayanan2017segnet} to largely reserve information about the source domain.
For transferring knowledge from the source agent to a new task, the environment reward is augmented with a shaped reward term to encourage the target agent to imitate the source agent on an embedded feature space:
\begin{align}
r'(s, \cdot) = \alpha\left\|f(s^s_{agent}; \theta_f) - g(s^t_{agent}; \theta_g) \right\|,
\end{align}
where $f(s^s_{agent})$ is the agent-specific state in the source domain, and $g(s^t_{agent})$ is for the target domain. 

Another work is \cite{ammar2012reinforcement} which applied the \textit{Unsupervised Manifold Alignment (UMA)} method~\cite{wang2009manifold} to automatically learn the {state mapping}.
%
Their approach requires collecting trajectories  from both the source and the target domain to learn such a mapping.
While applying policy gradient learning, trajectories from the target domain $\cM_t$ are first mapped back to the source: $\tau_t \to \tau_s$, 
then an expert policy in the source domain is applied to each initial state of those trajectories to generate near-optimal trajectories $\overset{\sim}{\tau_s}$, which are further mapped to the target domain: $\overset{\sim}{\tau_s} \to \overset{\sim}{\tau_t}$.
The deviation between $\overset{\sim}{\tau_t}$ and $\tau_t$ are used as a loss to be minimized in order to improve the target policy.
Similar ideas of using \textit{UMA} for inter-task mapping can also be found in~\cite{bocsi2013alignment} and~\cite{ammar2015unsupervised}.

In addition to approaches that utilizes mapping over states or actions, \cite{ammar2012ammarsparse} proposed to learn an inter-task mapping over the \textbf{\emph{transition dynamics}} space: $\cS \times \cA \times \cS$.
Their work assumes that the source and target domains are different in terms of the transition space dimensionality.
%
%
%
%
%
Transitions from both the source domain $\langle s^s, a^s, s'^s \rangle$ and the target domain $\langle s^t, a^t, s'^t \rangle$ are mapped to a latent space $Z$. 
Given the latent feature representations, a similarity measure can be applied to find a correspondence between the source and target task triplets.
Triplet pairs with the highest similarity in this feature space $Z$ are used to learn a mapping function $\cX$: $\langle s^t, a^t, s'^t \rangle = \cX( \langle s^s, a^s, s'^s \rangle)$.
%
After the transition mapping, states sampled from the expert policy in the source domain can be leveraged to render beneficial states in the target domain, which assists the target agent learning with a better initialization performance.
A similar idea of mapping transition dynamics can be found in~\cite{lazaric2008transfer}, which, however, requires a stronger assumption on the similarity of the transition probability and the state representations between the source and the target domains.

As summarized in Table~\ref{table:mapping}, for TL approaches that utilize an inter-task mapping, the mapped knowledge can be (a subset of) the state space~\cite{gupta2017learning,konidaris2006autonomous}, the $Q$-function~\cite{taylor2007transfer}, or (representations of) the state-action-sate transitions ~\cite{ammar2012ammarsparse}.
In addition to being directly applicable in the target domain~\cite{ammar2012ammarsparse}, the mapped representation can also be used as an augmented shaping reward~\cite{konidaris2006autonomous,gupta2017learning} or a loss objective~\cite{ammar2012reinforcement} in order to guide the agent learning in the target domain.
\judycom{
Most inter-task mapping methods tackle domains with moderate state-action space dimensions, such as maze tasks or tabular MDPs, where the goal can be reaching a target state with a minimal number of transitions.
Accordingly, \emph{tt} has been used to measure TL performance. 
For tasks with limited and discrete state-action space, evaluation is also conducted with different number of initial states collected in the target domain (\emph{nka}). 
}

\begin{table*}[htbp!]
  \centering
  \begin{tabular}{cclllc} 
    \multicolumn{1}{c}{\textbf{Methods}} & 
    \multicolumn{1}{l}{\begin{tabular}[c]{@{}c@{}} \textbf{RL} \\ \textbf{framework} \end{tabular}} & 
    \multicolumn{1}{l}{\begin{tabular}[c]{@{}c@{}} \textbf{MDP} \\ \textbf{difference} \end{tabular}} & 
    \multicolumn{1}{l}{\begin{tabular}[c]{@{}c@{}} \textbf{Mapping} \\ \textbf{function} \end{tabular}} & 
    %
    \multicolumn{1}{l}{\textbf{Usage of mapping}} & 
    %
    \multicolumn{1}{l}{\begin{tabular}[c]{@{}c@{}} \judycom{\textbf{Evaluation}} \\ \judycom{ \textbf{metrics}} \end{tabular}} \\ \hline
     {\cite{taylor2007transfer}} 
   & \multicolumn{1}{p{2cm}}{ \emph{SARSA} } 
   & \multicolumn{1}{p{3cm}}{ $\cS_t \neq \cS_t,\cA_s \neq \cA_t$} 
   & \multicolumn{1}{p{2.5cm}}{  $M(Q_s)$ $\to$ $Q_t$ } 
   & \multicolumn{1}{p{3cm}}{ $Q$ value reuse } 
   & {\emph{ \judycom{  ap, ar, tt, tr}} }
   \\ \hline
   {\cite{torrey2005using}} 
   & { \emph{Q-learning} } 
   & \multicolumn{1}{p{3cm}}{ $\cA_s \neq \cA_t$, $\cR_s \neq \cR_t$} 
   & \multicolumn{1}{p{3cm}}{  $M(Q_s)$ $\to$ \emph{advice} } 
   & \multicolumn{1}{p{3cm}}{  Relative $Q$ ranking}  
   & { \judycom{  \emph{ap, ar, tr}} }
   \\ \hline
   {\cite{gupta2017learning}} 
   & \multicolumn{1}{p{2cm}}{ $-$} 
   & \multicolumn{1}{p{3cm}}{ $\cS_s \neq \cS_t$} 
   & \multicolumn{1}{p{3cm}}{  $M(s_t)$ $\to$ $r'$ } 
   & \multicolumn{1}{p{3cm}}{ Reward shaping}  
   & { \judycom{  \emph{ap, ar, pe, tr}} }
   \\ \hline
   \cite{konidaris2006autonomous}
   &  \multicolumn{1}{p{2cm}}{ \emph{SARSA}$(\lambda)$ } 
   & \multicolumn{1}{p{3cm}}{ $\cS_s \neq \cS_t$  $\cR_s \neq \cR_t$} 
   & \multicolumn{1}{p{3cm}}{  $M(s_t)$ $\to$ $r'$ } 
   & \multicolumn{1}{p{3cm}}{ Reward shaping}  
   & { \judycom{  \emph{ap, ar, pe, tt}} }
   \\ \hline
  {\cite{ammar2012reinforcement}} 
   & \multicolumn{1}{p{2cm}}{ Fitted Value Iteration} 
   & \multicolumn{1}{p{2cm}}{ $\cS_s \neq \cS_t$} 
   & \multicolumn{1}{p{3cm}}{  $M(s_s)$ $\to$ $s_t$ } 
   & \multicolumn{1}{p{4cm}}{ Penalty loss on state deviation from expert policy}  
   & { \judycom{  \emph{ap, ar, pe, tr}} }
   \\ \hline
    {\cite{lazaric2008transfer}} 
   & \multicolumn{1}{p{2cm}}{ Fitted Q Iteration} 
   & \multicolumn{1}{p{3cm}}{ $\cS_s \times \cA_s \neq \cS_t \times \cA_t$} 
   & \multicolumn{1}{p{3cm}}{  $M \big ((s_s, a_s, s'_s)$ $\to$ $(s_t, a_t, s'_t) \big )$ } 
   & \multicolumn{1}{p{4cm}}{ Reduce random exploration }  
   & { \judycom{  \emph{ap, ar, pe, tr, nta}} }
   \\ \hline
     {\cite{ammar2012ammarsparse}} 
    & \multicolumn{1}{p{2cm}}{ $-$} 
    & \multicolumn{1}{p{3cm}}{ $\cS_s \times \cA_s \neq \cS_t \times \cA_t$} 
    & \multicolumn{1}{p{3cm}}{  $M \big ((s_s, a_s, s'_s)$ $\to$ $(s_t, a_t, s'_t) \big )$ } 
    & \multicolumn{1}{p{4cm}}{ Reduce random exploration }  
    & { \judycom{  \emph{ap, ar, pe, tr, nta}} }
    \\ \bottomrule 
  \end{tabular}
  \vspace{-0.05in}
  \caption{A comparison of \textit{inter-task mapping} approaches. \label{table:mapping} ``$-$'' indicates no RL framework constraints.}
  \label{tb:interTaskMapping}
  \end{table*}  
  \vspace{-0.1in}

\subsection{Representation Transfer} \label{sec:invariant}
This section review approaches that transfer knowledge in the form of representations learned by deep neural networks.
%
They are built upon the following consensual assumption:
\begin{assumption*}\emph{[Existence of Task-Invariance Subspace]}

    The state space ($\cS$), action space ($\cA$), or the reward space ($\cR$) can be disentangled into orthogonal subspaces, which are task-invariant 
    such that knowledge can be transferred between domains on the universal subspace.
\end{assumption*}

We organize recent work along this line into two subtopics: 1)  approaches that directly \emph{reuse} representations from the source domain (Section \ref{sec:reuse-rep}), and 2) approaches that {learn} to \emph{disentangle} the source domain representations into independent sub-feature representations, some of which are on the universal feature space shared by both the source and the target domains (Section \ref{sec:learnDisentangle}).

\subsubsection{Reusing Representations}\label{sec:reuse-rep}
A representative work of reusing representations is~\cite{rusu2016progressive}, which proposed the \emph{progressive neural network} structure to enable knowledge transfer across multiple RL tasks in a progressive way.
A progressive network is composed of multiple \emph{columns}, and each column is a policy network for one specific task.
It starts with one single column for training the first task, and then the number of columns increases with the number of new tasks.
While training on a new task, neuron weights on the previous columns are frozen, and representations from those frozen tasks are applied to the new column via a collateral connection to assist in learning the new task.
%

\modify{\emph{Progressive network} comes with a cost of large network structures, as the network grows proportionally with the number of incoming tasks.
A later framework called \emph{PathNet} alleviates this issue by learning a network with a fixed size~\cite{fernando2017pathnet}.}
\emph{PathNet} contains \emph{pathways}, which are subsets of neurons whose weights contain the knowledge of previous tasks and are frozen during training on new tasks.
The population of \emph{pathway} is evolved using a tournament selection genetic algorithm~\cite{harvey2009microbial}.

Another approach of reusing representations for TL is modular networks~\cite{devin2017learning, andreas2017modular, zhang2018decoupling}.
For example, \cite{devin2017learning} proposed to decompose the policy network into a task-specific module and agent-specific module.
%
%
Specifically, let  $\pi$ be a policy performed by any agent (robot) $r$ over the task $\cM_k$ as a function $\bm{\phi}$ over states $s$, it can be decomposed into two sub-modules $g_k$ and $f_r$, \ie:
\begin{align*}
    \pi(s) := \bm{\phi}(s_{env}, s_{agent}) = f_r(g_k(s_{env}), s_{agent}),
\end{align*}
where $f_r$ is the agent-specific module and $g_k$ is the task-specific module.
Their core idea is that the task-specific module can be applied to different agents performing the same task, which serves as a transferred knowledge. Accordingly, the agent-specific module can be applied to different tasks for the same agent.

A model-based approach along this line is \cite{zhang2018decoupling}, which learns a model to map the state observation $s$ to a latent-representation $z$.
The transition probability is modeled on the latent space instead of the original state space, \ie~$\hat{z}_{t+1}=f_{\theta}(z_t, a_t) $, where $\theta$ is the parameter of the transition model, $z_t$ is the latent-representation of the state observation, and $a_t$ is the action accompanying that state.
Next, a \emph{reward} module learns the value function as well as the policy from the latent space $z$ using an actor-critic framework.
One potential benefit of this latent representation is that knowledge can be transferred across tasks that have different rewards but share the same transition dynamics.
%
%
 
\subsubsection{Disentangling Representations} \label{sec:learnDisentangle} 
 
%
Methods discussed in this section mostly focus on learning a \emph{disentangled} representation.
Specifically, we elaborate on TL approaches that are derived from two techniques: {\emph{Successor Representation (SR)}} and {\emph{Universal Value Function Approximating (UVFA)}}.

\textit{\textbf{Successor Representations (SR)}} is an approach to decouple the state features of a domain from its reward distributions.
It enables knowledge transfer across multiple domains: $\bm{\cM} = \{\cM_1, \cM_2, \dots, \cM_K  \}$, so long as the only difference among them is the reward distributions: $\cR_{i} \neq \cR_j$.
\textit{SR} was originally derived from neuroscience,  until~\cite{dayan1993improving} proposed to leverage it as a generalization mechanism for state representations in the RL domain.
%

Different from the $v$-value or $Q$-value that describes states as dependent on the reward function, \textit{SR} features a state based on the \emph{occupancy measure} of its \emph{successor} states.
%
%
Specifically, \textit{SR} decomposes the value function of any policy into two independent components, $\vpsi$ and $R$:
    $V^\pi(s) = \sum_{s'} \vpsi(s, s') \vW(s')$, 
%
%
where $\vW(s')$ is a reward mapping function that maps states to scalar rewards, and $\vpsi$ is the \textit{SR} which describes any state $s$ as the occupancy measure of the future occurred states when following $\pi$, 
with $\cone[S=s']=1$ as an indicator function:
\begin{align*}
    \vpsi(s,s') = \cE_{\pi}[\sum_{i=t}^{\infty} \gamma^{i-t} \cone [S_{i}=s'] |S_t=s].
\end{align*}
\vspace{-0.1in}

The \emph{successor} nature of \textit{SR} makes it learnable using any TD-learning algorithms.
Especially, \cite{dayan1993improving} proved the feasibility of learning such representation in a tabular case, in which the state transitions can be described using a matrix.
\textit{SR} was later extended by~\cite{barreto2017successor} from three perspectives:
(i) the feature domain of \textit{SR} is extended from states to state-action pairs;
(ii) deep neural networks are used as function approximators to represent the \textit{SR} $\vpsi^\pi(s,a)$ and the  \emph{reward mapper} $\vW$;
(iii) Generalized policy improvement (GPI) algorithm is introduced to accelerate policy transfer for multi-tasks (Section~\ref{sec:GPI}). 
These extensions, however, are built upon a stronger assumption about the MDP:
\vspace{-0.05in}
\begin{assumption*}\emph{[Linearity of Reward Distributions]}
    The reward functions of all tasks can be computed as a linear combination of a fixed set of features:
        $r(s,a, s') = \phi(s,a,s')^{\top}\vW$,
    where $\phi(s,a,s') \in \RR^d$ denotes the latent representation of the state transition, and $\vW \in \RR^d$ is the task-specific reward mapper.
\end{assumption*}
\vspace{-0.05in}

Based on this assumption, \textit{SR} can be decoupled from the rewards when evaluating the Q-function of any policy $\pi$ in a task.
%
%
 The advantage of \textit{SR} is that, when the knowledge of $\vpsi^\pi(s,a)$ in the source domain $\cM_s$ is observed, one can quickly get the performance evaluation of the same policy in the target domain $\cM_t$
by replacing $\vW_s$ with $\vW_t$: $ Q^{\pi}_{\cM_t} = \vpsi^\pi(s,a) \vW_t.$
%
%
Similar ideas of learning \textit{SR} as a TD-algorithm on a latent representation $\phi(s,a,s')$ can also be found in \cite{kulkarni2016deep,zhang2017deep}.
Specifically, the work of \cite{kulkarni2016deep} was developed based on a weaker assumption about the reward function:
Instead of requiring linearly-decoupled rewards, the latent space $\phi(s,a,s')$ is learned in an encoder-decoder structure to ensure that the information loss is minimized when mapping states to the latent space.
This structure, therefore, comes with an extra cost of learning a decoder $f_d$ to reconstruct the state: $f_d( \phi(s_t)) \approx s_t$.

An intriguing question faced by the \textit{SR} approach is: \emph{Is there a way that evades the linearity assumption about reward functions and still enables learning the \textit{SR} without extra modular cost?}
An extended work of SR~\cite{barreto2019transfer} answered this question affirmatively, which proved that the reward functions does not necessarily have to follow the linear structure,
yet at the cost of a looser performance lower-bound while applying the GPI approach for policy improvement.
%
Especially, rather than learning a reward-agnostic latent feature $\phi(s,a,s') \in \RR^d$ for multiple tasks,
\cite{barreto2019transfer} aims to learn a matrix $\vphi(s,a,s') \in \RR^{D \times d}$ to interpret the basis functions of the latent space instead, where $D$ is the number of seen tasks.
Assuming $k$ out of $D$ tasks are linearly independent,  this matrix forms $k$ basis functions for the latent space. Therefore, for any unseen task $\cM_i$, its latent features can be built as a linear combination of these basis functions, as well as its reward functions $r_i(s,a,s')$. 
Based on the idea of basis-functions for a task's latent space, they proposed that $\vphi(s,a,s')$ can be approximated as learning $\RR(s,a,s')$ directly, where  $\RR(s,a,s') \in \RR^D$ is a vector of reward functions for each seen task: $$\RR(s,a, s') = \big [r_1(s,a,s'); r_2(s,a,s'), \dots, r_D(s,a,s') \big ].$$
Accordingly, learning $\vpsi(s,a)$ for any policy $\pi_i$ in $\cM_i$ becomes equivalent to learning a collection of Q-functions: 
%
$$ {\psi}^{\pi_i}(s,a)  = \big [Q^{\pi_i}_1(s,a), Q^{\pi_i}_2(s,a), \dots, Q^{\pi_i}_D(s,a) \big].$$ 
A similar idea of using reward functions as features to represent unseen tasks is also proposed by~\cite{mehta2008transfer}, which assumes the $\vpsi$ and $\vW$ as observable quantities from the environment.

\textbf{\textit{Universal Function Approximation} (UVFA)} is an alternative approach of learning disentangled state representations~\cite{schaul2015universal}.
Same as SR, \textit{UVFA}  allows TL for multiple tasks which differ only by their reward functions (goals).
Different from \textit{SR} which focuses on learning a reward-agnostic state representation, \textit{UVFA}  aims to find a function approximator that is generalized for both states and goals.
\modify{
The \textit{UVFA} framework is built on a specific problem setting of \emph{goal conditional RL}: 
\emph{task goals are defined in terms of states, \eg\ given the state space $\cS$ and the goal space $\cG$, it satisfies that $\cG \subseteq \cS$.}    
}
One instantiation of this problem setting can be an agent exploring different locations in a maze, where the goals are described as certain locations inside the maze.
Under this problem setting, a \textit{UVFA} module can be decoupled into a state embedding $\phi(s)$ and a goal embedding $\vpsi(g)$, by applying the technique of matrix factorization to a reward matrix describing the goal-conditional task.
%
%

One merit of \textit{UVFA} resides in its transferrable embedding $\phi(s)$ across tasks which only differ by goals.
Another benefit is its ability of continual learning when the set of goals keeps expanding over time.
On the other hand, a key challenge of \textit{UVFA} is that applying the matrix factorization is time-consuming, which makes it a practical concern for complex environments with large state space $|\cS|$.
Even with the learned embedding networks, the third stage of fine-tuning these networks via end-to-end training is still necessary.
%

\textit{UVFA} has been connected to \textit{SR} by \cite{barreto2019transfer}, in which a set of independent rewards (tasks) themselves can be used as features for state representations.
Another extended work that combines \textit{UVFA} with \textit{SR} is called \textit{Universal Successor Feature Approximator (USFA)}, which is proposed by~\cite{borsa2018universal}. 
Following the same linearity assumption, 
\textit{USFA} is proposed as a function over a triplet of the state, action, and a policy embedding $z$:
    $\phi(s,a, z): \cS \times \cA \times \RR^k \rightarrow \RR^d$,
%
where $z$ is the output of a \emph{policy-encoding mapping} $z = e(\pi): \cS \times \cA \rightarrow \RR^k$. 
Based on \textit{USFA}, the $Q$-function of any policy $\pi$ for a task specified by $\vW$ can be formularized as the product of a reward-agnostic \textit{Universal Successor Feature (USF)} $\vpsi$ and a reward mapper $\vW$:
$Q(s,a, \vW, z) =  \vpsi(s,a, z)^{\top} \vW.$
%
%
Facilitated by the disentangled rewards and policy generalization, \cite{borsa2018universal} further introduced a generalized TD-error as a function over tasks $\vW$ and policy $z$, which allows them to approximate the $Q$-function of any policy on any task using a TD-algorithm. 

\vspace{-0.05in}
\subsubsection{Summary and Discussion} \vspace{-0.05in}
We provide a summary of the discussed work in this section in Table \ref{tb:transfer-representation}.
Representation transfer can facilitate TL in multiple ways based on assumptions about certain task-invariant property.
Some assume that tasks are different only in terms of their reward distributions.
%
Other stronger assumptions include (i) decoupled dynamics, rewards~\cite{barreto2017successor}, or policies~\cite{borsa2018universal} from the $Q$-function representations, and (ii) the feasibility of defining tasks in terms of states~\cite{borsa2018universal}.
Based on those assumptions, approaches such as TD-algorithms~\cite{barreto2019transfer} or matrix-factorization~\cite{schaul2015universal} become applicable to learn such disentangled representations.
To further exploit the effectiveness of disentangled structure, we consider that \emph{generalization} approaches, which allow changing dynamics or state distributions, are important future work that is worth more attention in this domain.

\judycom{
Most discussed work in this section tackles multi-task RL or meta-RL scenarios, hence the agent's \emph{generalization} ability is extensively investigated.
For instance, methods of modular networks largely evaluated the \emph{zero-shot} performance from the meta-RL perspective~\cite{devin2017learning,borsa2018universal}.
Given a fixed number of training epochs (\emph{pe}),  \emph{Transfer ratio (tr)} is manifested differently among these methods.
It can be the relative performance of a modular net architecture compared with a baseline, or the accumulated return in modified target domains, where reward scores are negated for evaluating the dynamics transfer.
\emph{Performance sensitivity (ps)} is also broadly studied to estimate the \emph{robustness} of TL. 
\cite{barreto2017successor} analyzed the performance sensitivity given varying source tasks,
while \cite{borsa2018universal} studied the performance on different unseen target domains. 
}

There are unresolved questions in this intriguing research topic. 
One is \textbf{\emph{how to handle drastic changes of reward functions between domains}}. 
As discussed in~\cite{lehnert2017advantages}, good policies in one MDP may perform poorly in another due to the fact that beneficial states or actions in $\cM_s$ may become detrimental in $\cM_t$ with totally different reward functions. 
%
%
Learning a set of basis functions~\cite{barreto2019transfer} to represent unseen tasks (reward functions), or decoupling policies from $Q$-function representation~\cite{borsa2018universal} may serve as a good start to address this issue, as they propose a generalized latent space, from which different tasks (reward functions) can be interpreted.
However, the limitation of this work is that it is not clear how many and what kind of sub-tasks need to be learned to make the latent space generalizable enough.  

Another question is \emph{\textbf{how to generalize the representation learning for TL across domains with different dynamics or state-action spaces}}.
A learned \textit{SR} might not be transferrable to an MDP with different transition dynamics, as the distribution of occupancy measure for SR may no longer hold.
Potential solutions may include model-based approaches that approximate the dynamics directly or training a latent representation space for states using multiple tasks with different dynamics for better generalization~\cite{petangoda2019disentangled}.
Alternatively, TL mechanisms from the supervised learning domain, such as meta-learning, which enables the ability of fast adaptation to new tasks~\cite{finn2017model}, or importance sampling~\cite{zadrozny2004learning}, which can compensate for the prior distribution changes~\cite{pan2009survey}, may also shed light on this question.

\begin{table*}[htbp!]
    \centering
    \begin{tabular}{cccccc}  
      \multicolumn{1}{c}{\textbf{Methods}} &    
      \multicolumn{1}{c}{\textbf{Representations format}} &
       \multicolumn{1}{c}{\textbf{Assumptions }} & 
       \multicolumn{1}{p{2.5cm}}{\textbf{MDP difference}} & 
       \multicolumn{1}{c}{\textbf{Learner}}  &
    \multicolumn{1}{l}{\begin{tabular}[c]{@{}c@{}} \judycom{\textbf{Evaluation}} \\ \judycom{ \textbf{metrics}} \end{tabular}} \\ \hline
     \multicolumn{1}{c}{ Progressive Net \cite{rusu2016progressive}}  
     & \multicolumn{1}{p{3cm}}{Lateral connections to previously learned network modules } 
     & \multicolumn{1}{c}{ N/A} 
     & {  $\cS, \cA$}
     & \multicolumn{1}{c}{  A3C}
     & \multicolumn{1}{c}{  \judycom{  $ap,ar, pe, ps, tr$}} \\ \hline
     \multicolumn{1}{c}{ PathNet \cite{fernando2017pathnet}}  
     & \multicolumn{1}{p{3cm}}{Selected neural paths } 
     & \multicolumn{1}{c}{ N/A }
     & \multicolumn{1}{c}{  $\cS, \cA$}
     & \multicolumn{1}{c}{  A3C}
     & \multicolumn{1}{c}{ \judycom{  $ap, ar, pe, tr$}} \\ \hline
     \multicolumn{1}{c}{ Modular Net \cite{devin2017learning}}  
     & \multicolumn{1}{p{3cm}}{Task(agent)-specific network module} 
     & \multicolumn{1}{p{3cm}}{Disentangled state representation}
     & \multicolumn{1}{c}{  $\cS, \cA$}
     & \multicolumn{1}{c}{  Policy \newline Gradient}
     & \multicolumn{1}{c}{ \judycom{  $ap, ar, pe, tt$}} 
     \\ \hline
     \multicolumn{1}{c}{Modular Net \cite{zhang2018decoupling}}  
     & \multicolumn{1}{p{3cm}}{Dynamic transitions module learned on state latent representations.} 
     & { N/A}
     & \multicolumn{1}{c}{  $\cS, \cA$}
     & \multicolumn{1}{c}{  A3C }
     & \multicolumn{1}{c}{ \judycom{  $ap,ar, pe, tr, ps$}} \\ \hline
     \multicolumn{1}{c}{SR  \cite{barreto2017successor}}  
     & \multicolumn{1}{l}{\textit{SF}} 
     & \multicolumn{1}{p{3cm}}{Reward function can be linearly decoupled}
     & \multicolumn{1}{c}{  $\cR$}
     & \multicolumn{1}{c}{  DQN }
     & \multicolumn{1}{c}{ \judycom{  $ap,ar, nka, ps$}} \\ \hline
     \multicolumn{1}{c}{ SR  \cite{kulkarni2016deep}}  
     & \multicolumn{1}{p{3cm}}{Encoder-decoder learned \textit{SF}} 
     & \multicolumn{1}{c}{N/A}
     & \multicolumn{1}{c}{  $\cR$}
     & \multicolumn{1}{c}{  DQN }
     & \multicolumn{1}{c}{\judycom{   $ap, ar, pe, ps$}} \\ \hline
     \multicolumn{1}{c}{SR \cite{barreto2019transfer}}  
     & \multicolumn{1}{p{3cm}}{Encoder-decoder learned \textit{SF}} 
     & \multicolumn{1}{p{3cm}}{Rewards can be represented by set of basis functions}
     & \multicolumn{1}{c}{  $\cR$}
     & \multicolumn{1}{c}{  $Q(\lambda)$ }
     & \multicolumn{1}{c}{  \judycom{  $ap, pe$}} \\ \hline
     \multicolumn{1}{c}{UVFA \cite{schaul2015universal}}  
     & \multicolumn{1}{p{3cm}}{Matrix-factorized UF} 
     & \multicolumn{1}{p{3cm}}{Goal conditional RL}
     & \multicolumn{1}{c}{  $\cR$}
     & \multicolumn{1}{c}{  Tabular  Q-learning}
     & \multicolumn{1}{c}{ \judycom{   $ap, ar, pe, ps$}} \\ \hline
     \multicolumn{1}{c}{UVFA with SR \cite{borsa2018universal}}  
     & \multicolumn{1}{p{3cm}}{Policy-encoded UF} 
     & \multicolumn{1}{p{3cm}}{Reward function can be linearly decoupled}
     & \multicolumn{1}{c}{  $\cR$}
     & \multicolumn{1}{c}{  $\epsilon$-greedy Q-learning}
     & \multicolumn{1}{c}{ \judycom{  $ap, ar, pe$}} \\ \bottomrule
    \end{tabular}
    \vspace{-0.1in}
    \caption{A comparison of TL approaches of \textit{representation transfer}.\label{tb:transfer-representation}}
    \vspace{-0.1in}
    \end{table*}   
\vspace{-0.15in}
\section{Applications} \label{sec:applications} \vspace{-0.05in}
In this section we summarize recent applications that are closely related to using TL techniques for tackling RL domains. 

\modify{
\textbf{\emph{Robotics learning}} is a prominent application domain of RL.
%
TL approaches in this field include \emph{robotics learning from demonstrations}, where expert demonstrations from humans or other robots are leveraged~\cite{argall2009survey}
Another is \emph{collaborative robotic training}~\cite{kehoe2015survey,gu2017deep}, in which knowledge from different robots is transferred by sharing their policies and episodic demonstrations.
%
Recent research focus is this domain is fast and robust adaptation to unseen tasks.
%
One example towards this goal is \cite{yu2017preparing}, in which robust robotics policies are trained using synthetic demonstrations to handle dynamic environments.
Another solution is to learn domain-invariant latent representations.
Examples include~\cite{sadeghi2016cad2rl}, which learns the latent representation using 3D CAD models, and~\cite{bousmalis2018using,bharadhwaj2019data} which are derived based on the Generative-Adversarial Network.
Another example is \emph{DARLA}~\cite{higgins2017darla}, which is a zero-shot transfer approach to learn disentangled representations that are robust against domain shifts.
We refer readers to \cite{zhao2020sim,kober2013reinforcement} for detailed surveys along this direction. 

\textbf{\emph{Game Playing}} is a common test-bed for TL and RL algorithms. It has evolved from classical benchmarks such as grid-world games to more complex settings such as online-strategy games or video games with multimodal inputs.
One example is \emph{AlphaGo}, which is an algorithm for learning the online chessboard games using both TL and RL techniques~\cite{silver2016mastering}. 
\emph{AlphaGo} is first pre-trained offline using expert demonstrations and then learns to optimize its policy using Monte-Carlo Tree Search. Its successor, \emph{AlphaGo Master}~\cite{silver2017mastering}, even beat the world's first ranked human player. 
TL-DRL approaches are also thriving in video game playing.
Especially, OpenAI has trained \emph{Dota2} agents that can surpass human experts~\cite{OpenAI2019blog}.
State-of-the-art platforms include \emph{MineCraft}, \emph{Atari}, and \emph{Starcraft}. 
\cite{oh2016control} designed new RL benchmarks under the \emph{MineCraft} platform.
%
%
\cite{justesen2019deep} provided a comprehensive survey on DL applications in video game playing, which also covers TL and RL strategies from certain perspectives.
A large portion of TL approaches reviewed in this survey have been applied to the \emph{Atari} platforms ~\cite{mnih2013playing}. 

\textbf{\emph{Natural Language Processing (NLP)}} has evolved rapidly along with the advancement of DL and RL.
%
Applications of RL to NLP range widely, from \emph{Question Answering (QA)}~\cite{chen2017survey}, \emph{Dialogue systems}~\cite{singh2000reinforcement}, \emph{Machine Translation}~\cite{zoph2016neural}, to an integration of NLP and Computer Vision tasks, such as \emph{Visual Question Answering (VQA)}~\cite{hu2017learning}, \emph{Image Caption}~\cite{ren2017deep}, etc.
Many NLP applications have implicitly applied TL approaches.
%
Examples include learning from expert demonstrations for \emph{Spoken Dialogue Systems}~\cite{andreas2016learning}, \emph{VQA}~\cite{hu2017learning}; or reward shaping for \emph{Sequence Generation}~\cite{bahdanau2016actor}, \emph{Spoken Dialogue Systems}~\cite{su2015reward},\emph{QA}~\cite{lin2018multi,godin2019learning}, and \emph{Image Caption}~\cite{ren2017deep}, or transferring policies for \emph{Structured Prediction}~\cite{chang2015learning} and \emph{VQA}~\cite{lu2017best}.
}

\vspace{-0.05in}
\judycom{
\emph{\textbf{Large Model Training}}:
   RL from human and model-assisted feedback becomes indispensable in training 
   large models (LMM), such as GPT4~\cite{openai2023gpt}, Sparrow~\cite{glaese2022improving}, PaLM~\cite{chowdhery2022palm}, LaMDA~\cite{thoppilan2022lamda},
   which have shown tremendous breakthrough in dialogue applications, search engine answer improvement, artwork generation, etc.
   The TL method at the core of them is using human preferences as a reward signal for model fine-tuning, where the preference ranking itself is considered as shaped rewards. 
   We believe that TL with carefully crafted human knowledge will help better align large models with human intent and hence  achieve trustworthy and de-biased AI.
}

\textbf{\emph{Health Informatics}}:
RL has been applied to various healthcare tasks~\cite{yu2019reinforcement}, including 
\emph{automatic medical diagnosis}~\cite{alansary2019evaluating,ma2017multimodal},
\emph{health resource scheduling}~\cite{gomes2017reinforcement}, and \emph{drug discovery and development},~\cite{serrano2018accelerating,popova2018deep}, etc.
%
%
Among these applications we observe emerging trends of leveraging prior knowledge to improve the RL procedure, especially given the difficulty of accessing large amounts of clinical data.
Specifically, \cite{gaweda2005incorporating} utilized $Q$-learning for drug delivery individualization. They integrated the prior knowledge of the dose-response characteristics into their $Q$-learning framework to avoid random exploration.
%
%
%
\cite{killian2017robust} applied a DQN framework for prescribing effective HIV treatments, in which they learned a latent representation to estimate the uncertainty when transferring a pertained policy to the unseen domains.
\cite{holzinger2016interactive} studied applying human-involved interactive RL training for health informatics.

\emph{\textbf{Others}}: RL has also been utilized in many other real-life applications.
Applications in the \textbf{\emph{Transportation Systems}} have adopted RL to address traffic congestion issues with better \emph{traffic signal scheduling} and \emph{transportation resource allocation}~\cite{wei2018intellilight,li2016traffic,el2013multiagent,lin2018efficient}. We refer readers to \cite{yau2017survey} for a review along this line.
Deep RL are also effective solutions to problems in \textbf{\emph{Finance}}, including \emph{portfolio management}~\cite{moody1998performance,jiang2017cryptocurrency}, \emph{asset allocation}~\cite{neuneier1998enhancing}, and \emph{trading optimization}~\cite{deng2016deep}.
Another application is the \textbf{\emph{Electricity Systems}}, especially the \emph{intelligent electricity networks}, which can benefit from RL techniques for improved power-delivery decisions~\cite{dalal2016hierarchical,ruelens2016residential} and active resource management~\cite{wen2015optimal}. \cite{glavic2017reinforcement} provides a detailed survey of RL techniques for electric power system applications.
%

\section{Future Perspectives} \label{sec:futurePerspectives}
\judycom{In this section, we present some open challenges and future directions in TL that are closely related to the DRL domain, based on both retrospectives of the methods discussed in this survey and outlooks to the emerging trends of AI.}

\vspace{0.05in}
\judycom{
\noindent \textbf{\emph{Transfer Learning from Black-Box:}}
}
\judycom{
%
%
Ranging from exterior teacher demonstrations to pre-trained function approximators, black-box resource is more accessible and predominant than well-articulated knowledge.
Therefore, leveraging such black-box resource  is indispensable for practical TL in DRL.
One of its main challenges resides in \textbf{\emph{estimating the optimality}} of black-box resource, which can be potentially noisy or biased.  
We consider that efforts can be made from the following perspectives:
\begin{enumerate}
\item Inferring the \emph{reasoning} mechanism inside the black-box. Resemblant ideas have been explored in \emph{inverse RL} and \emph{model-based RL}, where the goal is to approximate the reward function or to learn the dynamics model under which the demonstrated knowledge becomes reasonable.
 
\item Designing effective \emph{feedback} schemes, including leveraging domain-provided rewards, intrinsic reward feedback, or using human preference as feedback.

\item Improving the \emph{{interpretability}} of the transferred knowledge~\cite{li2017infogail,ramakrishnan2016towards}, which benefits in evaluating and explaining the process of TL from black-box. 
It can also alleviate catastrophic decision-making for high-stake tasks such as auto-driving.
\end{enumerate}
}

\vspace{0.05in}
\judycom{
\noindent \textbf{\emph{Knowledge Disentanglement and Fusion}}
are both towards better knowledge sharing across domains.
Disentangling knowledge is usually a \emph{prerequisite} for efficient knowledge fusion, which may involve external knowledge from multiple source \emph{domains}, with diverging \emph{qualities}, presented in different \emph{modalities}, etc.
Disentangling knowledge in RL can be interpreted from different perspectives: i) disentangling the action, state, or reward representations, as discussed in Sec \ref{sec:invariant}; 2) decomposing complex tasks into multiple \emph{skill snippets.}
The former is an effective direction in tackling meta-RL and multi-task RL, although some solutions hinge on strict assumptions of the problem setting, such as linear dependence among domain dynamics or learning goals.
The latter is relevant to hierarchical RL and \emph{prototype learning} from sequence data~\cite{choi2016retain}.
It is relatively less discussed besides few pioneer research \cite{petangoda2019disentangled}. We believe that this direction is worth more research efforts, which not only benefits interpretable knowledge learning, but also  aligns with human perception.
}

\vspace{0.05in} 
\modify{
\noindent \textbf{\emph{Framework-Agnostic Knowledge Transfer}}: Most contemporary TL approaches are designed for certain RL frameworks. Some are applicable to RL algorithms designed for the discrete-action space, while others may only be feasible given a continuous action space. One fundamental reason behind is the diversified development of RL algorithms. We expect that unified RL frameworks would contribute to the standardization of TL approaches in this field. 
}

\vspace{0.05in} 
\modify{
\noindent \textbf{\emph{Evaluation and Benchmarking}}:
Variant evaluation metrics have been proposed to measure TL from different but complementary perspectives, although no single metric can summarize the efficacy of a TL approach. Designing a set of generalized, novel metrics is beneficial for the development of TL in DRL. 
Moreover, with the effervescent development of large-scale models, it is crucial to standardize evaluation from the perspectives of \emph{ethics} and \emph{groundedness}. 
The \emph{appropriateness} of the transferred knowledge, such as potential \emph{stereotypes} in human preference, and the \emph{bias} in the model itself should also be quantified as metrics. 
}

\vspace{0.05in}
\judycom{
\noindent \textbf{\emph{Knowledge Transfer to and from Pre-Trained Large Models:}}
By the time of this survey being finalized, unprecedented DL breakthroughs have been achieved in learning large-scale models built on massive computation resource and attributed data.
One representative example is the Generative Pre-trained Transformer (GPT)~\cite{openai2023gpt}.
%
%
Considering them as complete \textbf{\emph{knowledge graphs}} whose training process maybe inaccessible, there are more challenges in this direction besides learning from a \emph{black-box}, which are faced by a larger AI community including the RL domain. 
We briefly point out two directions that are worth ongoing attention:
\begin{enumerate}
    \item \emph{Efficient model fine-tuning with knowledge distillation.} One important method for fine-tuning large models is \emph{{RL with human feedback}}, in which the quantity and quality of human ratings are critical for realizing a good reward model.
    We anticipate  other forms of TL methods in RL to be explored to further improve the efficiency of  fine-tuning, such as imitation learning with adversarial training to achieve human-level performance.
    
    \item \emph{Principled prompt-engineering for knowledge extraction}. More often the large model itself cannot be accessed, but only input and output of models are allowed.
    Such inference based knowledge extraction requires delicate prompt designs. Some efficacious efforts include designing prompts with mini task examples as {one-shot learning},  \emph{decomposing}  complex tasks into \emph{architectural}, {contextual} prompts.
    Prompt engineering is being proved an important direction for effective knowledge extraction, which with proper design, can largely benefit downstream tasks that depend on large model resources.
\end{enumerate}
}
\vspace{-0.15in}
\section{Acknowledgements}\vspace{-0.05in}
This research was supported by the National Science Foundation (IIS-2212174, IIS-1749940), National Institute of Aging (IRF1AG072449), and the Office of Naval Research (N00014-20-1-2382).
\vspace{-0.05in}

\vspace{-0.1in}
\bibliography{references}
\bibliographystyle{IEEEtran}

\vspace{-0.3in}
\vspace{-0.1in}
\begin{IEEEbiography}[{\includegraphics[width=1in,height=1.25in,clip,keepaspectratio]{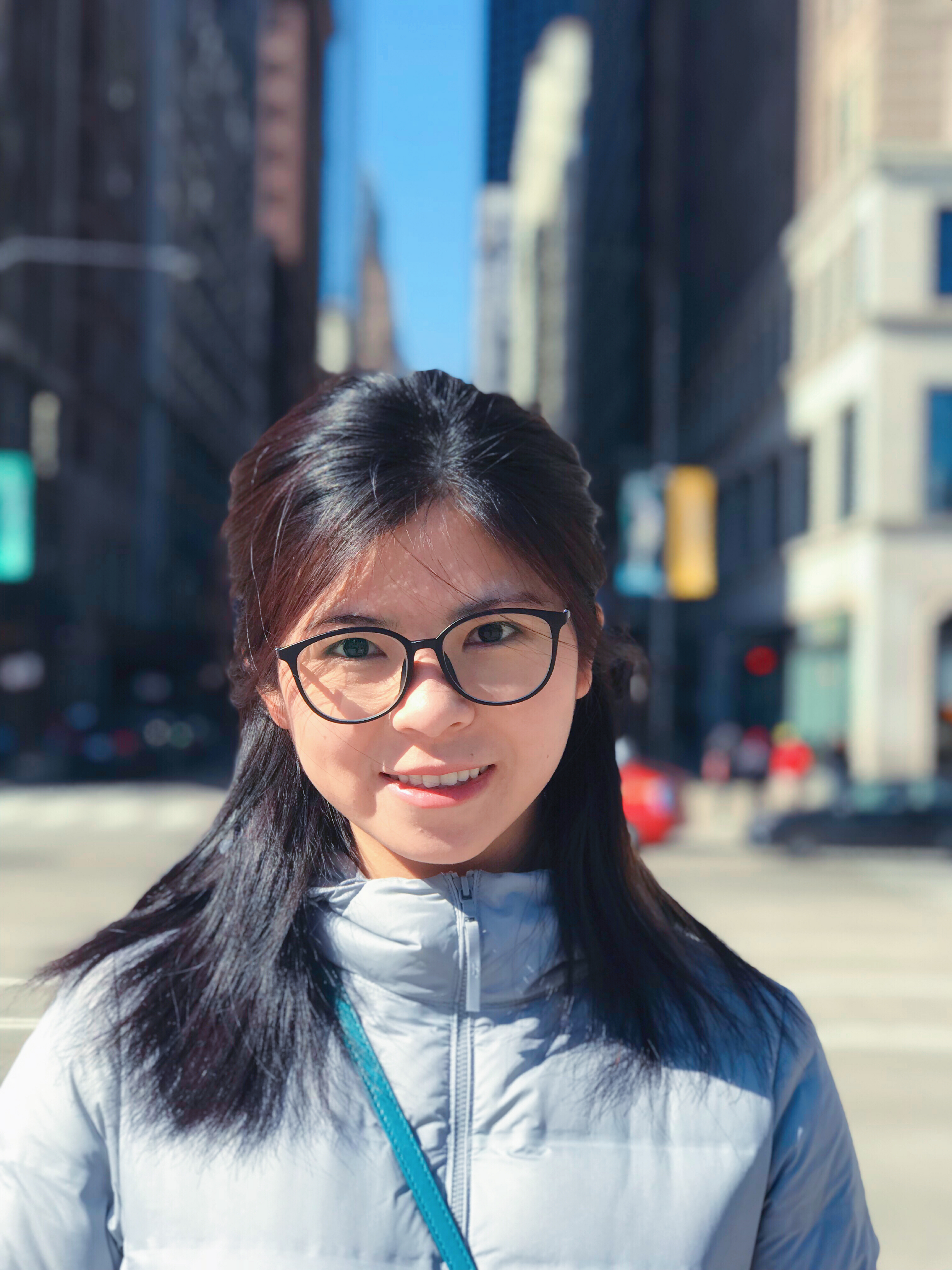}}]{Zhuangdi Zhu} is currently a senior data and applied scientist with Microsoft. She obtained her Ph.D. degree from the Computer Science department of Michigan State University. Zhuangdi has regularly published on prestigious machine learning conferences including NeurIPs, ICML, KDD, AAAI, etc. 
Her research interests reside in both fundamental and applied machine learning. 
Her current research involves reinforcement learning and distributed machine learning.
\end{IEEEbiography}

\vspace{-0.3in}
\vspace{-0.1in}
\vspace{-0.15in}
\begin{IEEEbiography}[{\includegraphics[width=1in,height=1.25in,clip,keepaspectratio]{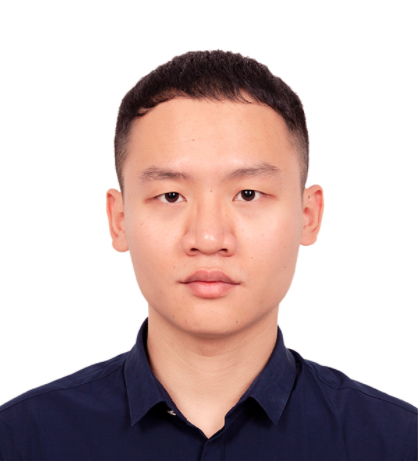}}]{Kaixiang Lin} is  an applied scientist at Amazon web services. He obtained his Ph.D. from Michigan State University. He has broad research interests across multiple fields, including reinforcement learning, human-robot interactions, and natural language processing. His research has been published on multiple top-tiered machine learning and data mining conferences such as ICLR, KDD, NeurIPS, etc. He serves as a reviewer for top machine learning conferences regularly. 
\end{IEEEbiography}

\vspace{-0.3in} 
\vspace{-0.1in}
\vspace{-0.15in}

\begin{IEEEbiography}[{\includegraphics[width=1in,height=1.25in,clip,keepaspectratio]{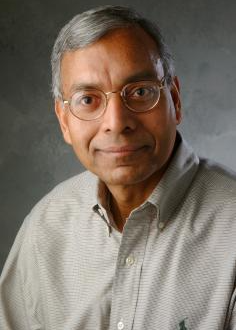}}]{Anil K. Jain }
        is a University distinguished professor in the Department of Computer Science and Engineering at Michigan State University. His research interests include pattern recognition and biometric authentication. He served as the editor-in-chief of the IEEE Transactions on Pattern Analysis and Machine Intelligence and was a member of the United States Defense Science Board. He has received Fulbright, Guggenheim, Alexander von Humboldt, and IAPR King Sun Fu awards. He is a member of the National Academy of Engineering and a foreign fellow of the Indian National Academy of Engineering and the Chinese Academy of Sciences.
\end{IEEEbiography}

\vspace{-0.3in}
\vspace{-0.1in}
\vspace{-0.05in}

\begin{IEEEbiography}[{\includegraphics[width=1in,height=1.25in,clip,keepaspectratio]{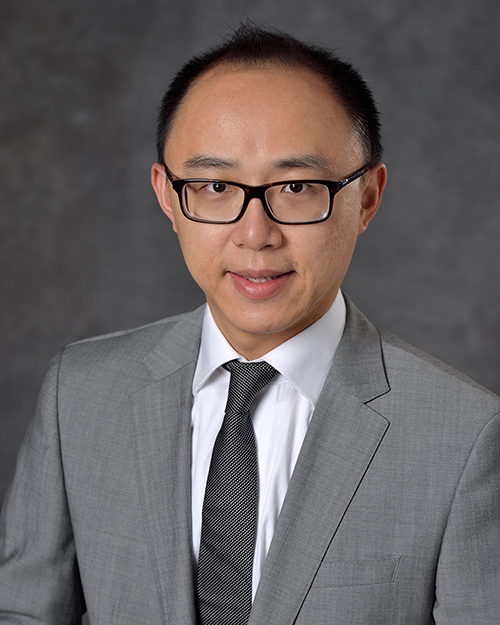}}]{Jiayu Zhou} is an associate
        professor in the Department of Computer Science and Engineering at Michigan State University. He received his Ph.D. degree in computer science from Arizona State University in 2014. He has broad research interests in the fields of large-scale machine learning and data mining as
        well as biomedical informatics. He has served as a technical program committee member for
        premier conferences such as NIPS, ICML, and SIGKDD. His papers have received the Best Student Paper Award at the 2014 IEEE International Conference on Data Mining (ICDM), the Best Student Paper Award at the 2016 International Symposium on Biomedical Imaging (ISBI) and the Best Paper Award at IEEE Big Data 2016.
\end{IEEEbiography}

\end{document}